\documentclass[nohyperref]{article}

\usepackage{microtype}
\usepackage{graphicx}
\usepackage{subfigure}
\usepackage{booktabs} 

\usepackage{hyperref}

\usepackage[accepted]{icml2022}

\usepackage{amsmath}
\usepackage{amssymb}
\usepackage{amsfonts}
\usepackage{mathtools}
\usepackage{amsthm}
\usepackage{dsfont}

\DeclareMathOperator*{\argmax}{arg\,max} 

\usepackage[capitalize,noabbrev]{cleveref}

\theoremstyle{plain}

\theoremstyle{definition}

\theoremstyle{remark}

\usepackage[textsize=tiny]{todonotes}

\icmltitlerunning{Safe-RL with Minimal Supervision}

\begin{document}

\twocolumn[
\icmltitle{Safe Reinforcement Learning with Minimal Supervision}




\begin{icmlauthorlist}
\icmlauthor{Alexander Quessy}{bris}
\icmlauthor{Thomas Richardson}{bris}
\icmlauthor{Sebastian East}{bris}
\end{icmlauthorlist}

\icmlaffiliation{bris}{Department of Aerospace Engineering, University of Bristol, Bristol, United Kingdom}

\icmlcorrespondingauthor{Alexander Quessy}{aq15777@bristol.ac.uk}

\icmlkeywords{Safe Reinforcement Learning, Unsupervised Learning}

\vskip 0.3in
]



\printAffiliationsAndNotice{}  

\begin{abstract}

Reinforcement learning (RL) in the real world necessitates the development of procedures that enable agents to explore without causing harm to themselves or others. The most successful solutions to the problem of safe RL leverage offline data to learn a safe-set, enabling safe online exploration. However, this approach to safe-learning is often constrained by the demonstrations that are available for learning.

In this paper we investigate the influence of the quantity and quality of data used to train the initial safe learning problem offline on the ability to learn safe-RL policies online. Specifically, we focus on tasks with spatially extended goal states where we have few or no demonstrations available. Classically this problem is addressed either by using hand-designed controllers to generate data or by collecting user-generated demonstrations. However, these methods are often expensive and do not scale to more complex tasks and environments. To address this limitation we propose an unsupervised RL-based offline data collection procedure, to learn complex and scalable policies without the need for hand-designed controllers or user demonstrations. Our research demonstrates the significance of providing sufficient demonstrations for agents to learn optimal safe-RL policies online, and as a result, we propose optimistic forgetting, a novel online safe-RL approach that is practical for scenarios with limited data. Further, our unsupervised data collection approach highlights the need to balance diversity and optimality for safe online exploration.


\end{abstract}

\section{Introduction}
\label{sec:introduction}
Reinforcement learning (RL) is a powerful technique and has enabled robotic systems to complete a variety of tasks \citep{ICML1, ICML2, ICML3} in high-dimensional environments \citep{ICML4} under uncertainty. Typically, modern RL agents learn through extensive unconstrained interaction with their environment, aiming to maximize a reward based objective function. However, unconstrained exploration is hazardous in the real-world, posing harm to both the agent and its environment. This presents a fundamental challenge to the real-world use of RL: understanding the effects of our actions to guarantee safety whilst not knowing the consequences of unsafe transitions until they are explored \citep{ICML35}.

Critically, we need to infer how our agent's actions will impact its behavior before we deploy it into the real-world. Planning with LMPC \citep{ICML9} offers a useful approach to predict the effect of a controller's actions without explicitly sampling a possibly dangerous state. Generally, LMPC methods rely upon dense reward functions, to learn policies efficiently, but complex real-world tasks rarely offer an intuitive representation of task success with a simple dense reward function representation. Instead, we often perceive complex tasks as problems with temporally and/or spatially extended goal-states with a binary success criteria. For example an aerial robot landed or crashed. Directly encoding complex tasks with a simple binary reward function upon reaching the goal state offers little supervision during the learning process, degrading sample efficiency and requiring greater exploration. Often reward shaping \citep{ICML5} is used to address this problem, improving the density of the reward function at the cost of generalization and scalability. However, reward shaping masks the true problem, which is that fundamentally we need to provide more information to the agent if we want to learn efficiently, but naturally this will be task specific. This is often evident from the ineffectiveness of reward shaping in tasks with complex state-spaces, such as image based observations \citep{ICML6, ICML7, ICML8}: they are simply too complicated for the designer to provide useful information through the reward function.

A logical next step, when the task is too complex to learn from reward shaping alone, is to learn from demonstrations in the form of offline reinforcement learning \citep{ICML34}. The combination of offline RL, safe-sets, and LMPC presents a promising method for safe-RL \citep{ICML32, ICML10}. This approach enables learning a sub-optimal control policy that can be refined online, with exploration restricted by the safe-set \citep{ICML9}. To learn an effective safe online policy we require a dataset containing:

\begin{itemize}
    \item \textbf{Goal reaching demonstrations}, where the agent reaches the goal, to learn how to reach the goal state.
    \item \textbf{Constraint violating demonstrations}, where the agent intentionally violates the task constraints to learn the bounds of the safe-set.
\end{itemize}

In previous work demonstrations were generated using: discrete hand-designed task specific controllers \citep{ICML11, ICML12}, trained online stochastic controllers \citep{ICML28, ICML29}: using reward shaping, or user demonstrations \citep{ICML30, ICML31}. Whilst effective, all these approaches require task specific information provided by the practitioner to generate useful demonstrations, limiting their ability to scale and generalize to new tasks and environments.

An alternative, previously unexplored, approach to this safe-RL problem is to use unsupervised RL \cite{ICML17} to automatically generate a task specific dataset. A technique that does not rely upon the practitioner to design a task specific controller. For safe-RL this is a problem of exploration, gathering sufficient demonstrations to learn a complex representation online and classification, encoding a representation of goal-reaching and constraint violating behavior. In this work we focus on which elements of offline learning are critical for safe online learning and how to effectively make use of unsupervised RL for offline RL. We make the following contributions:
\begin{itemize}
    \item In section \ref{sec:few_demos} we analyze the impact of data \textit{quantity} on training using demonstrations from a discrete controller on a complex goal-oriented visual motor task. We then offer an improvement called \textit{optimistic forgetting} which helps to mitigate the binary failure mode that can occur when learning from few offline demonstrations online.
    \item In section \ref{sec:no_teacher} we analyze the impact of data \textit{quality} on training making use of unsupervised learning to generate datasets automatically without the need for a hand-designed controller. This allows us to learn generalizable safe-RL policies, agnostic of task specific objectives. We found one of the largest challenges in this area is working out how to balance data quality over quantity, as it is easy to generate plenty of data with unsupervised learning, but providing a balance between sub-optimal demonstrations and constraint violating behavior is important for successful adaption online. 
\end{itemize}

\section{Related Work}
\label{sec:related_work}
\subsection{Optimizing for Performance and Safety}

\citet{ICML9} first introduced Learning Model Predictive Control (LMPC) with the aim to iteratively improve controller performance from an initial feasible trajectory within a reference free setting. They found by iteratively learning a value function and safe-set from task roll-outs, theoretic guarantees can be made concerning stability, optimality and solution feasibility for non-linear or stochastic linear systems \citep{ICML14, ICML15}. \citet{ICML12} extend LMPC to higher order visuomotor tasks and LS$^{3}$ \citet{ICML11} improved upon this by optimizing a task specific cost function, rather than uniformly expanding across the state-space. We build upon this work, investigating tasks in higher order state-spaces, offering algorithmic improvements that prove critical when learning from few offline demonstrations.

\subsection{Learning Safe Behaviors from Demonstrations}

Using demonstrations to learn safe-sets offline has proven to be an effective method to safely constraint exploration in online RL. In contrast to previous approaches where exploration is restricted by hard constraints \citet{ICML47} or a safety critic \citet{ICML48}, our approach restricts exploration with a safe-set. Previous LMPC based approaches include LS$^{3}$ \citep{ICML11} and SAVED \citep{ICML12} where they aim to predict the likelihood of constraint violation or leaving the safe-set. Similarly, \citet{ICML16} Recovery RL queries actions over a task-policy distribution and is used with model-free or model-based RL methods.

All of these previous approaches obtained demonstrations either from hand-designed controllers or user demonstrations. We aim to understand the impact of these demonstrations for online learning. We explore the importance of controlling implicit bias within the dataset used for offline learning, and how quantity and diversity of demonstrations effects task-specific performance \citep{ICML36}.

\subsection{Unsupervised Offline Learning}
\citet{ICML26} and \citet{ICML27} argue decoupling exploration and exploitation into independent separate components should help to improve scalability of RL; recent work by \citet{ICML24, ICML25} in offline learning provide evidence of this. Similar arguments can be made for safe-RL and to some extent the offline to online learning setup already separates out exploration and exploitation. In safe-RL the problem is more complicated as we need to not only explore the environment to find suitable behaviors, but classify them under constraint-violating or goal-reaching categories. \citet{ICML17} provide a useful comparison of different unsupervised approaches using the DeepMind Control Suite \citep{ICML37}. They found there is still significant algorithmic work to do in order to reach state-of-the-art performance with unsupervised learning and that there is a large gap in performance when using state and image based observations. In the unsupervised numerical experiments presented in this paper, we focused on using state-based observations because the results are easier to interpret than image-based observations. We also found competence based approaches \citep{ICML21, ICML22, ICML23} useful due to their expressive latent skill vector encoding allowing for behavioral classification. It is important to note that these unsupervised RL methods are relatively weak \citep{ICML17}, and we expect that as advancements are made in this area, the procedures outlined in this work will also improve.

\section{Problem Statement}
\label{sec:problem_statement}
We consider episodic finite time-horizon goal-conditioned tasks, characterized by a Constrained Markov Decision Process (CMDP) and described with a tuple: $\mathcal{M} := ( \mathcal{S}, \mathcal{G}, \mathcal{A}, P, R, C, \mu, T )$.

\begin{itemize}
    \item $\mathcal{S}$, $\mathcal{G}$ and $\mathcal{A}$ are the state, goal and action spaces respectively. We consider image states with pixel width $W$ and height $H$ and direct state representations where $\mathcal{S} = \mathbb{R}^{n}$. The state variable at time $t$ is denoted $s_{t} \in \mathcal{S}$.
    \item $P(s_{t+1}|s_{t}, a_{t})$ maps states and actions to a probability distribution over subsequent states, $P: \mathcal{S} \times \mathcal{A} \times \mathcal{S} \rightarrow \mathbb{R}$.
    \item $R(s_{t+1}, s_{t}, a_{t})$ is the CMDP reward function, $R: \mathcal{S} \times \mathcal{A} \times \mathcal{S} \rightarrow \mathbb{R}$.
    \item $C(s_{t})$ is a constraint indicator function, $C : \mathcal{S} \to \{0, 1\}$, $1$ if violated.
    \item $\mu$ is the initial state distribution ($s_{0} \sim \mu$) and $T$, the time horizon.
\end{itemize}

The agent's objective is to reach a goal state $\mathcal{G} \subset \mathcal{S}$ in the fewest number of steps whilst avoiding task specific constraints. To represent the difficulty of reward shaping in real-world finite horizon tasks, similar to \citet{ICML12} and \citet{ICML11}, we use a discrete reward:

$$
    R(s, a, s') =\begin{cases}
        0 & \text{$s' \in \mathcal{G}$} \\
        -1, & \text{$s' \notin \mathcal{G}$}.
    \end{cases}
$$

For all tasks, the goal state was defined as a ball with radius $\epsilon$ centered on a single state value, $s_{g} \in \mathcal{S}$:

$$ 
    \mathcal{G} := \{s : \| s - s_{g} \| \leq \epsilon \}.
$$


Given a policy $\pi : \mathcal{S} \rightarrow \mathcal{A}$, we define total return over a CMDP $\mathcal{M}$ as $R^{\pi} = \mathbb{E}_{\pi, \mu, P} [\sum_{t=0}^{T}R(s_{t}, a_{t})]$. Similar to \citet{ICML11}, we define $P_{\mathcal{C}}^{\pi}$ as the probability of future constraint violation over the CMDP time-horizon under policy $\pi$ from state $s$. The objective then becomes to maximize the expected return $R^{\pi}$, whilst maintaining a constraint violation probability less than $\delta_{\mathcal{C}}$, equation \ref{eq:prob_viol}:

\begin{equation}
    \label{eq:prob_viol}
    \begin{aligned}
    \pi^{*} = \argmax_{\pi \in \Pi} \quad & \mathbb{E}_{\pi, \mu, P} \Bigl[ \sum_{t=0}^{T}R(s_{t}, a_{t}) \Bigr] \\
    \textrm{s.t.} \quad & \mathbb{E}_{s_{0} \sim \mu} [P_{\mathcal{C}}^{\pi}(s_{0})] \leq \delta_{\mathcal{C}}.
    \end{aligned}
\end{equation}

\section{Safe Learning from Few Demonstrations}
\label{sec:few_demos}

Obtaining high quality goal-reaching demonstrations can be challenging and expensive, requiring demonstrations from task/domain experts or for the practitioner to spend time hand-designing a controller. To understand the influence of data quantity on online safe-RL we conducted experiments using three navigation tasks with varying complexity, and a range of dataset sizes. The following navigation tasks were investigated and are explained in detail in appendix \ref{sec:te_and_c}:

\begin{itemize}
    \item \textbf{SimplePointBot} (SPB), the agent needs to navigate around a block, by controlling its velocity.
    \item \textbf{Bottleneck}, the agent needs to navigate down a pipe between 2 chambers without hitting the walls, by controlling its velocity.
    \item \textbf{SimpleVelocityBot} (SVB), the agent needs to navigate around a block, by controlling its acceleration.
\end{itemize}

To learn an optimal policy online we use an LMPC approach similar to those outlined in LS$^{3}$ \citep{ICML11} and SAVED \citep{ICML12}. To learn from the RGB $64\times64$ pixel image observations used in this section, we employed a $\beta$-variational auto-encoder, $f_{enc}$ and $f_{dec}$, as described in \citet{ICML39}. The encoder, $f_{enc}$ transforms states from the dataset $\mathcal{D}$ into a $d$-dimensional latent space $\mathcal{Z}$, whilst the decoder, $f_{dec}$, reconstructs the original images. We then used a probabilistic dynamics model, similar to PETS \citep{ICML38}, to estimate the likelihood of the agent: reaching the goal, violating constraints, or generating high reward. These likelihoods are estimated by querying learned deep neural network based representations for the safe-set $f_\mathcal{S}$, value-function $V_{\pi}$, goal-indicator $f_{\mathcal{G}}$ and constraint estimator $f_{\mathcal{C}}$. This training procedure is described in more detail in Appendix \ref{sec:safeRL}.

\begin{figure}[htb!]
    \centering
    \includegraphics[width=1.0\columnwidth]{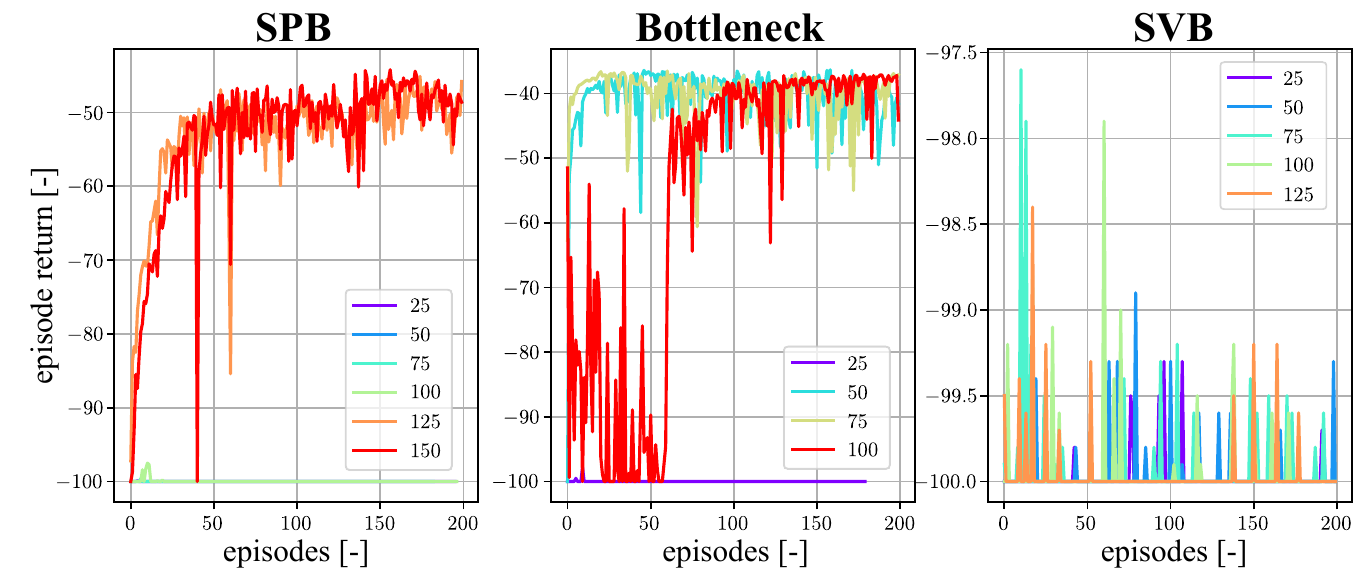}
    \caption{Episodic return from the 3 navigation environments using the LMPC procedure outlined in LS$^{3}$. SPB requires a minimum of 125 demonstrations to learn a goal-reaching policy and SVB is unable to learn a goal-reaching policy at all.}
    \label{fig:A-demo_compare}
\end{figure}

When the quantity of demonstrations used to train the LMPC approach outlined above is too low the agent is unable to learn a goal-reaching policy online. As depicted in figure \ref{fig:A-demo_compare}, when fewer than 125 demonstrations are used, SPB experiences a failure to generate any reward, and when fewer than 50 demonstrations are used, Bottleneck encounters the same issue. Additionally, SVB is unable to learn a reliable policy online using any number of demonstrations. The presence of this binary-failure mode online highlights the importance of using a sufficient quantity of demonstrations to train safe-RL agents.

\begin{figure}[htb!]
    \centering
    \includegraphics[width=1.0\columnwidth]{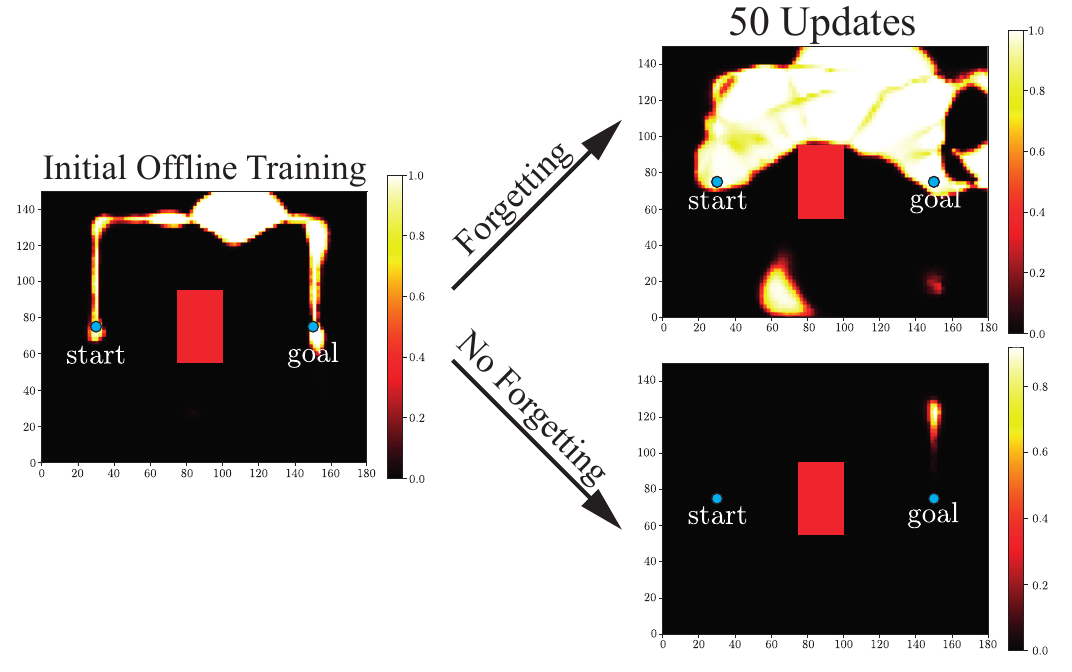}
    \caption{Heatmap of the Safe-Set $f_{\mathcal{S}}(s)$ for SPB. Left: initial offline training. Top right: after 50 updates with optimistic forgetting. Bottom right: after 50 updates without optimistic forgetting.}
    \label{fig:B-safeset}
\end{figure}

We discovered the reason for this binary-failure mode is that when too few demonstrations are provided, the agent frequently fails to reach the goal state because the learned policy is not robust enough to reach the goal state. This leads to the value function and safe-set shrinking in size during online training, resulting in the cost of exploration becoming excessively high and the agent rarely venturing beyond the initial starting state. As shown in figure \ref{fig:B-safeset}, after only 50 demonstrations, the safe-set is effectively null as the agent expects all transitions to be unsafe, causing it to remain at the initial starting position.


To address this failure mode, we propose a modification to the LMPC procedure described above called \textit{optimistic forgetting}. This approach involves dropping the last $N_{\text{forget}}$ episodes from the replay buffer $\mathcal{D}$ if the normalized return $R$ is less than the minimum return $R_{\text{min}}$ over the past $N_{\text{forget}}$ episodes. Similar to accept-reject sampling \citep{ICML33}, this technique reinforces confidence in goal-reaching behavior by keeping the safe-set and goal-indicator open to exploration towards the goal, while also improving the accuracy of the learned dynamics, value function and constraint estimator, effectively introducing an optimism bias \citep{ICML41}. This approach assumes that the dataset used for offline learning includes some demonstrations of goal-reaching behavior. Algorithm \ref{alg:safe-set} describes the safe-learning procedure used in this work based on LS$^{3}$ with optimistic forgetting shown in \textcolor{orange}{orange}.

\begin{algorithm}[tb]
    \caption{Safe-Set LMPC w/ Optimistic Forgetting}
    \label{alg:safe-set}
 \begin{algorithmic}
    \STATE {\bfseries Input:} offline dataset $\mathcal{D}$, number of episodes $N_{\text{episodes}}$, \textcolor{orange}{forgetting frequency $N_{\text{forget}}$, minimum return $R_{\text{min}}$}
    \IF{state $s \in \mathcal{D}$ of type \textit{images}}
        \STATE Train: VAE encoder $f_{enc}$ and decoder $f_{dec}$ using data from $\mathcal{D}$
    \ENDIF
    \STATE Train: dynamics $f_{dyn}$, safe-set classifier $f_{\mathcal{S}}$, value function $V_{\pi}$, goal indicator $f_{G}$ and constraint estimator $f_{\mathcal{C}}$ using data from $\mathcal{D}$
    \STATE Current Return $R = 0$
    \FOR{$i_{\text{episode}} = 0$ {\bfseries to} $N_{\text{\text{episodes}}}$}
        \STATE Initialize environment $\mathcal{M}$ with $s_{0} \sim \mu$
        \FOR{$t = 0$ {\bfseries to} $T_{\text{episode}}$}
            \STATE Select and execute $a_{t}$ in $\mathcal{M}$ with PETS MPC
            \STATE Observe state $s_{t}$, reward $r_{t}$, constraint $c_{t}$
            \STATE $\mathcal{D} := \mathcal{D} \cup \{ (s_{t}, a_{t}, s_{t+1}, r_{t}, c_{t}) \}$
            \textcolor{orange}{\STATE $R \mathrel{+}= r_{t}$}
        \ENDFOR
        \STATE Update: $f_{dyn}$, $f_{\mathcal{S}}$, $V_{\pi}$, $f_{G}$, $f_{\mathcal{C}}$
        \textcolor{orange}{
            \IF{$i_{\text{episode}} \mathbin{\%} N_{\text{forget}} \equiv 0 \cup R \leq R_{\text{min}}$}
                \STATE Drop $\mathcal{D}_{i_{\text{episode}} - N_{\text{forget}}} \subset \mathcal{D}$
                \STATE $R=0$
            \ENDIF
        }
    \ENDFOR
\end{algorithmic}
\end{algorithm}

\begin{figure}[htb!]
    \centering
    \includegraphics[width=1.0\columnwidth]{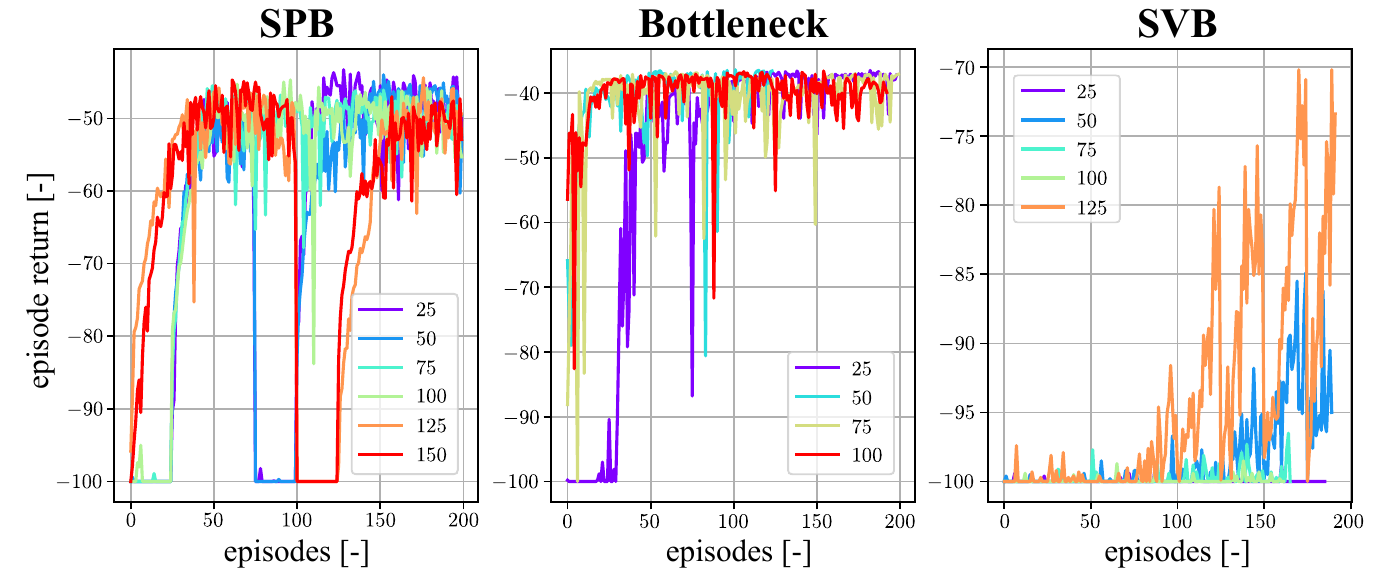}
    \caption{Episodic return from the 3 navigation environments, trained using the procedure outlined in algorithm \ref{alg:safe-set}, using optimistic forgetting. For SPB we can learn an optimal policy from as few as 25 demonstrations and SVB is able to learn a goal-reaching policy with 125 demonstrations.}
    \label{fig:C-optimistic}
\end{figure}

The improvement provided by optimistic forgetting is illustrated in Figure \ref{fig:C-optimistic}, which shows the results of the same experiments depicted in Figure \ref{fig:A-demo_compare}, but with the inclusion of the optimistic forgetting technique. Notably SPB can now learn from as few as 25 demonstrations and SVB is able to learn a goal-reaching policy. Figure \ref{fig:B-safeset} demonstrates how optimistic forgetting allows the safe-set to grow during online training and helps to overcome the binary failure mode present in the original setup. In these experiments we set the $N_{\text{forget}}$ term to 25 episodes and normalized $R_{\text{min}}$ to $0.5$. Performance could likely be improved by experimenting with these parameters further.

These empirical results emphasize that the quantity of demonstrations used for learning is a key factor for safe exploration online. In the next section we expand upon this by considering the impact of data quality on online exploration, specifically the significance of data diversity.

\section{Safe Learning without a Teacher}
\label{sec:no_teacher}

Designing controllers capable of generating useful goal-reaching and constraint-violating behaviors becomes increasingly difficult and time-consuming as task complexity grows. Unsupervised learning is an attractive alternative to hand designed controllers. It enables the automatic generation of diverse datasets without the need for extensive engineering effort, providing the practitioner with large amounts of data. Unsupervised RL is usually framed as a 2 stage process of:

\begin{enumerate}
    \item \textbf{Pretraining} a policy based on an intrinsic reward function to generate a diverse group of behaviors.
    \item \textbf{Fine-tuning} the policy to maximize a task specific extrinsic reward function.
\end{enumerate}

Fundamentally unsupervised RL is a method of finding useful behaviors without relying on a traditional reward function, whilst also ensuring a balance between diversity and utility. Safe-RL adds a layer of complexity, requiring behaviors to also be classified as either constraint-violating or goal-reaching in order to generate samples to train the LMPC procedure outlined in algorithm \ref{alg:safe-set}. 

Competence based unsupervised RL algorithms express the policy using a prior encoded vector representation, denoted as $\pi(s|z)$, where the skill vector is sampled from a distribution $z \sim \mathcal{Z}$. This prior-encoded skill vector offers a convenient representation for classifying behavior types and subsequently sampling from the policy.

To investigate the effectiveness of unsupervised learning and the subsequent effects of balancing optimality and diversity, we considered three experimental setups. Each one uses the SimplePointBot environment with state-based observations:

\begin{itemize}
    \item \textbf{Controller}: all demonstrations come from the deterministic controllers outlined in Appendix \ref{sec:te_and_c}.
    \item \textbf{Semi-Supervised}: constraint-violating demonstrations come from the deterministic controller and goal-reaching demonstrations from SAC, trained using the $p^{*}$ objective in equation \ref{eq:pstar}.
    \item \textbf{Unsupervised}: all demonstrations come from the unsupervised learning procedure described in algorithm \ref{alg:unsupervised}.
\end{itemize}

Soft-Actor Critic (SAC) \citep{ICML19} a state-based off-policy RL algorithm is used to train the semi-supervised and unsupervised policies used in the last 2 experiments. SAC's objective is to learn a policy that not only maximizes the expected discounted sum of rewards, but also maximizes the expected entropy of the policy over the policy state distribution $\rho_{\pi}(s)$:

\begin{equation}
    J(\pi) = \sum^{T}_{t=0} \mathbb{E}_{(\mathbf{s}_{t}, \mathbf{a}_{t}) \sim \rho_{\pi}(\mathbf{s}_{t})} [r(\mathbf{s}_{t}, \mathbf{a}_{t}) + \alpha \mathcal{H}(\pi(\cdot | \mathbf{s}_{t})) ]
    \label{eq:sac-reward}
\end{equation}

The balance between maximizing the expected discounted sum of rewards and maximizing the expected entropy of the policy is controlled by the $\alpha$ parameter. In this work optimization is completed using the procedure outlined in \citet{ICML19}, and a state-representation with a prior $\mathbf{s}_{t} = (\mathbf{o}_{t};\mathbf{z}_{t})$ and intrinsic reward $r_{\mathbf{z}_{t}}(\mathbf{s}_{t})$. As detailed in Appendix \ref{sec:SMM} we also considered using DrQ-v2 \citep{ICML20}, similar to URLB, but found it challenging to manage the explicit exploration schedule implemented in DrQ-v2.

State-Marginal Matching (SMM) \citep{ICML22}, a competence based unsupervised RL algorithm is used to pretrain the prior skill-encoded policy and generate goal-reaching and constraint-violating samples. Based on DIAYN \citep{ICML21}, SMM includes a $p^{*}(s)$ target distribution term in the intrinsic reward function, which significantly improves the efficiency of learning goal-reaching behaviors. Additionally, SMM's discrete one-hot vector encoding facilitates easy classification and sampling of demonstrations from each prior, a challenge faced by algorithms like APS \citep{ICML23}.

SMM combines the entropy based mutual information criteria found in DIAYN (Appendix \ref{sec:DIAYN}):
$$
    \log q_{\phi}(z|s) - \log p(z)
$$
with a state-marginal matching objective:
$$
    \min\limits_{\pi \in \Pi} D_{KL} (\rho_{\pi}(s) || p^{*}(s))
$$
These objectives can be combined as an intrinsic reward (Appendix \ref{sec:SMM}) and used in algorithm \ref{alg:unsupervised}:
\begin{equation}
    \label{eq:smm-reward}
    r_{z}(s) = \log p^{*} (s) - \log \rho_{\pi_{z}} (s) + \log p(z|s) - \log p(z)
\end{equation}

In order to make the data-collection process more efficient, we use a target distribution, $p^{*}(s)$, based on the distance from the goal state, $g$. This is defined as:

\begin{equation}
    \label{eq:pstar}
    p^{*}(s) = \sqrt{||s_{t} - g_{t}||_{2}^{2}}
\end{equation}

In experimentation, we found it necessary to incorporate a penalty for violating the constraint, along with a bonus for attaining the goal-state, in order to prevent the agent from getting trapped on the constraint block.

\begin{algorithm}[htb!]
    \caption{Unsupervised Safe Dataset Collection}
    \label{alg:unsupervised}
\begin{algorithmic}
\STATE {\bfseries Input:} environment $\mathcal{M}$, intrinsic reward $r_{intrinsic}$, discount factor $\gamma$, pretrain-steps $N_{pt}$, exploratory-samples $N_{samples}$, dataset-size $N_{dataset}$, prior-distribution $z \sim \mathcal{Z}$, minimum return $R_{min}$
\STATE {\bfseries Randomly Initialize:} actor $\pi_{\theta}$, critic $Q_{\phi}$
\FOR{$i_{step} = 0$ {\bfseries to} $N_{pt}$}
    \STATE Sample skill $z \sim p(z)$, initial state $s_{0} \sim \mu(s)$
    \FOR{$t = 0$ {\bfseries to} $T_{episode}$}
        \STATE $a_{t} \sim \pi_{\theta}(a_{t} | s_{t}, z)$
        \STATE $s_{t+1} \sim P(s_{t+1}|s_{t}, a_{t})$
        \STATE Calculate $r_{z}(s_{t})$ using equation \ref{eq:smm-reward}
        \STATE $\mathcal{D}_{replay} := \mathcal{D}_{replay} \cup \{ s_{t}, a_{t}, r_{t}, s_{t+1}, z_{t}\}$
        \STATE Update $\pi_{\theta}$, $Q_{\phi_{1, 2}}$, using mini-batches from $\mathcal{D}_{replay}$ with SAC
    \ENDFOR
\ENDFOR
\FOR{$i_{z} = 0$ {\bfseries to} $N_{samples}$}
    \STATE Sample skill $z \sim p(z)$, initial state $s_{0} \sim \mu(s)$
    \STATE Collect trajectory $\tau_{i_{z}} = \{ (s_{t}, a_{t}, s_{t+1}, r_{t}, c_{t}) \}$ with $\pi_{\theta}(.|., z)$ from $\mathcal{M}$
    \IF{$\sum_{t=0}^{T} r_{t} \geq R_{min}$}
        \STATE Store prior $z \cup Z_{gr}$
    \ENDIF
\ENDFOR
\FOR{$i_{dataset} = 0$ {\bfseries to} $N_{dataset}$}
    \STATE Sample skill $z \sim Z_{gr}$, initial state $s_{0} \sim \mu(s)$
    \STATE Collect trajectory $\tau_{i_{z}} = \{ (s_{t}, a_{t}, s_{t+1}, r_{t}, c_{t}) \}$ with $\pi_{\theta}(.|., z)$ from $\mathcal{M}$
    \STATE Store trajectory $\mathcal{D}_{gr} := \mathcal{D}_{gr} \cup \tau$
\ENDFOR
\STATE $i_{data} = 0$
\WHILE{$i_{data} < N_{dataset}$}
    \FOR{$i_{z} = 0$ {\bfseries to} $N_{samples}$}
        \STATE Sample skill $z \sim p(z)$, initial state $s_{0} \sim U(s)$
        \STATE Collect trajectory $\tau_{i_{z}} = \{ (s_{t}, a_{t}, s_{t+1}, r_{t}, c_{t}) \}$ with $\pi_{\theta}(.|., z)$ from $\mathcal{M}$
        \IF{$c_{t} = True \in \tau$}
            \STATE Store trajectory $\mathcal{D}_{constr} := \mathcal{D}_{constr} \cup \tau$
        \ENDIF
    \ENDFOR
    \STATE $i_{data} \mathrel{+}= 1$
\ENDWHILE
\STATE {\bfseries Return} $\mathcal{D} = \mathcal{D}_{constr} \cup \mathcal{D}_{gr}$
\end{algorithmic}
\end{algorithm}

Algorithm \ref{alg:unsupervised} describes the unsupervised RL approach used to generate data without the use of hand-designed controllers. Compared to classical unsupervised RL \citep{ICML17} we do not fine-tune the policy on a task specific reward function because the goal-oriented task setup has too sparse a reward function to update the policy. In our approach, goal-reaching behavior is obtained directly from the unsupervised prior, identified by searching for episodes with priors that generate extrinsic rewards from the goal state above a predefined bound $R_{min}$:

\begin{equation}
    \pi(s|z_{gr}) \rightarrow \sum_{t=0}^{T} r_{t} \geq R_{min}.
\end{equation}

We can then sample from those priors exclusively, to generate a dataset of goal-reaching behavior $\mathcal{D}_{gr}$. Constraint violating behaviors simply require sufficient demonstrations to learn the constraint predictor along with the underlying dynamics. These are gathered by selecting behaviors that violate constraints when sampling over a random starting state from all possible states in the environment (i.e. $\mu \sim \mathcal{U}(s)$). 

\begin{figure}[htb!]
    \centering
    \includegraphics[width=0.8\columnwidth]{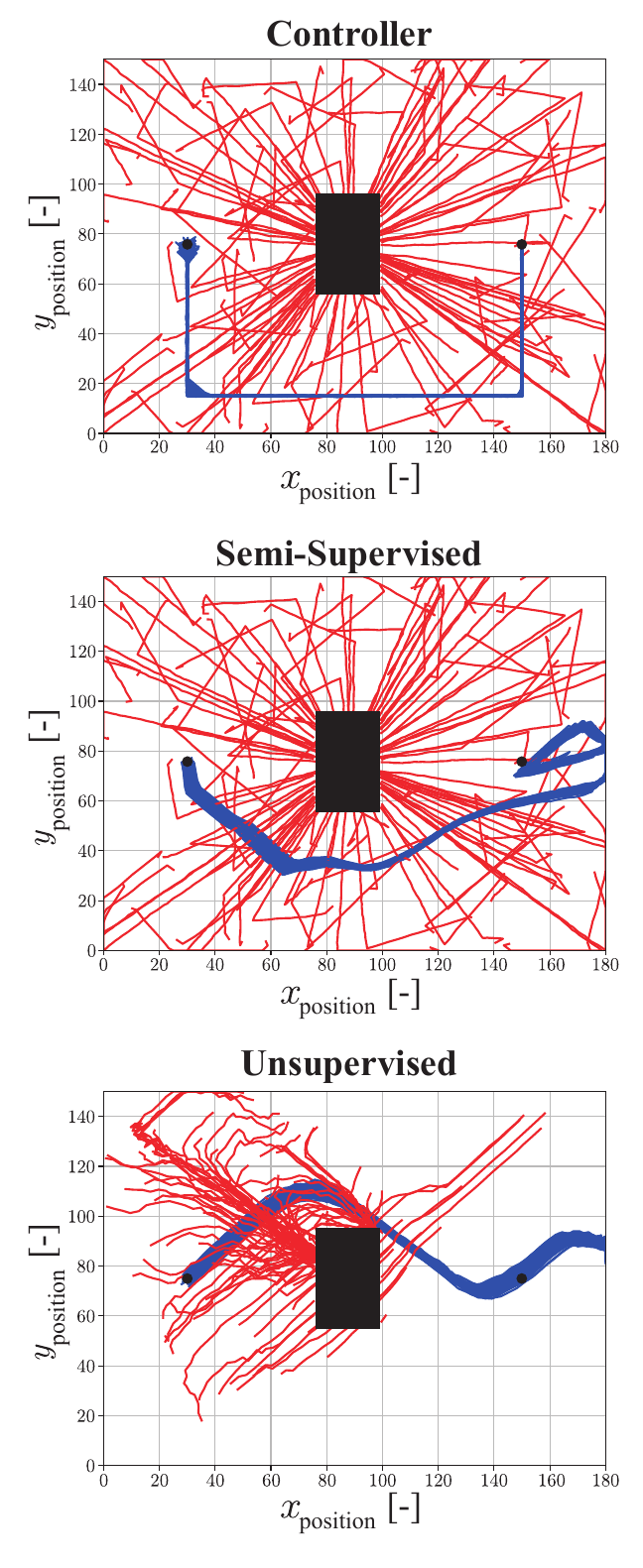}
    \caption{Unsupervised demonstrations for Safe-RL experiments. Constraint-violating trajectories are shown in \textcolor{red}{red} and goal-reaching trajectories in \textcolor{blue}{blue}.}
    \label{fig:D-unsupervised_demos}
\end{figure}


The agent was trained for a total of 4 million steps using an NVIDIA RTX8000 GPU for both semi-supervised and unsupervised experiments. A 4-dimensional one-hot skill vector encoding was used for $\mathcal{Z}$ in the SMM algorithm, with goal-reaching demonstrations generated by sampling from one of these priors. All 4 priors were found to be effective in producing constraint-violating demonstrations when starting from a random initial state. In both experiments the goal-reaching behavior successfully navigated to the goal position, but encountered difficulties in halting once the goal was reached. Figure \ref{fig:D-unsupervised_demos} illustrates this behavior, showing the dataset used in our experiments, composed of 200 demonstrations split evenly between goal-reaching and constraint-violating behaviors.

\begin{figure}[htb!]
    \label{fig:F-online_safe_RL}
    \centering
    \includegraphics[width=0.9\columnwidth]{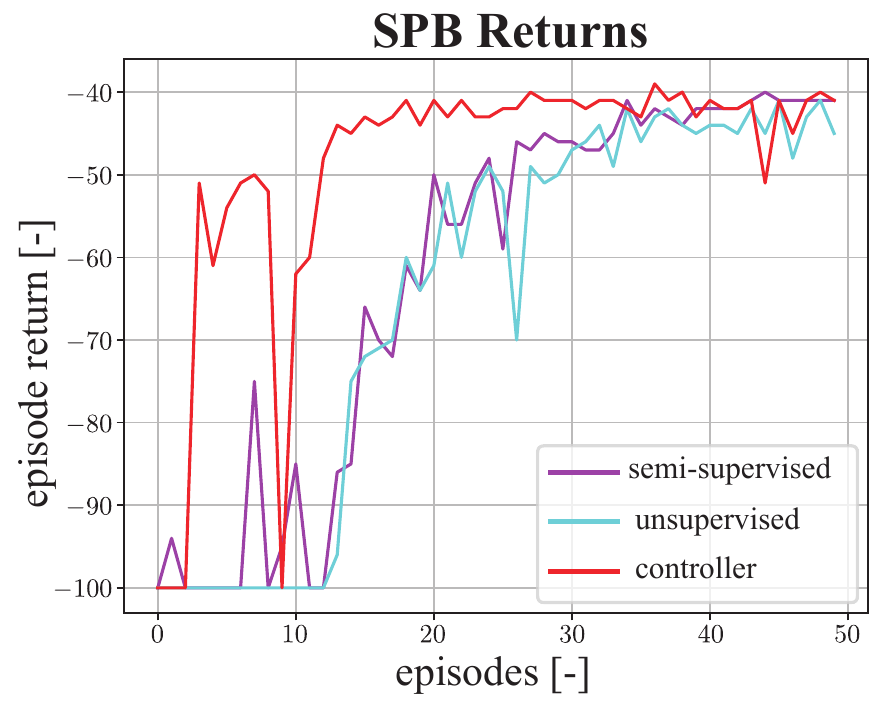}
    \caption{Episodic return using the LMPC procedure outlined in algorithm \ref{alg:safe-set} after initially training offline using the datasets depicted in figure \ref{fig:D-unsupervised_demos}.}
\end{figure}

\begin{figure}[htb!]
    \label{fig:E-offline_learning}
    \centering
    \includegraphics[width=1.0\columnwidth]{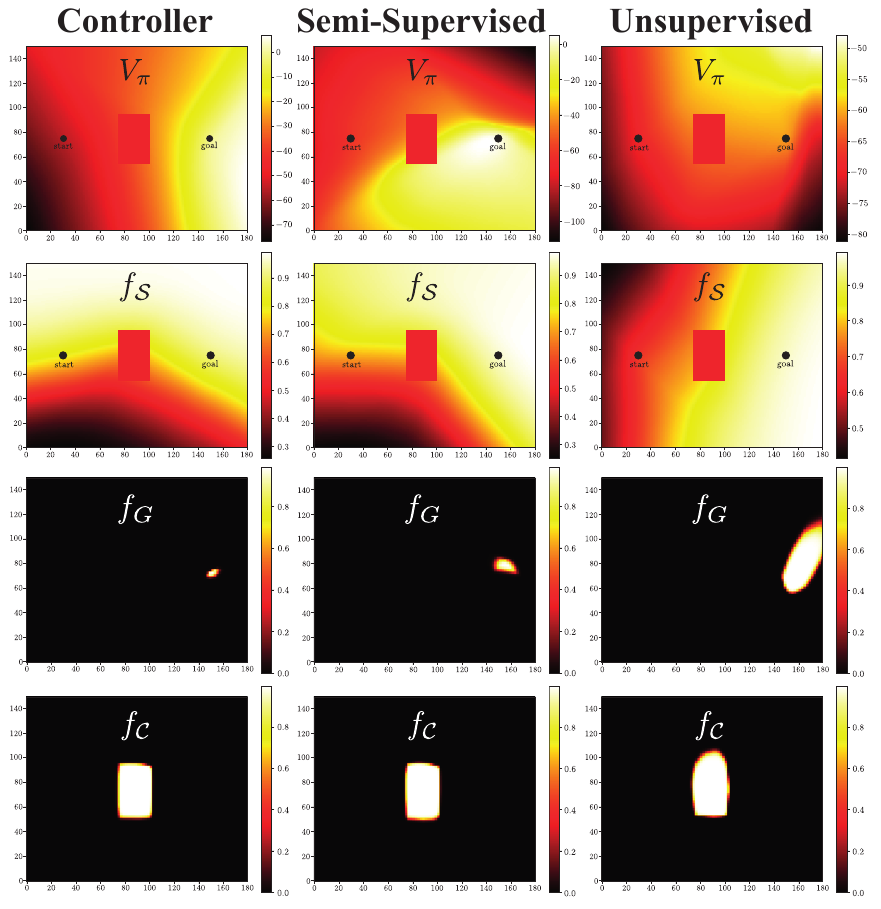}
    \caption{Offline learned: value $V_{\pi}(s)$, constraint $f_{\mathcal{C}}(s)$, safe-set $f_{\mathcal{S}(s)}$ and goal-indicator $f_{\mathcal{G}}(s)$ heatmaps for each set of demonstrations.}
\end{figure}

Figure \ref{fig:F-online_safe_RL} compares the online episodic return after training the agent on each of the 200 demonstration datasets used in the experiments. We found that the semi-supervised and unsupervised procedures took longer to learn a goal-reaching policy online than the discrete controller and both approaches required more episodes to learn a robust goal-reaching policy and subsequently converge to an optimal policy. The heatmaps in Figure \ref{fig:E-offline_learning}, show the offline learning progress for each of the key safe-RL functions learned offline. In the semi-supervised and unsupervised experiments the value function is not as tightly centered on the goal state when compared to the controller experiment. This explains why the agent took longer to learn an optimal policy, and is likely caused by the agent's inability to immediately halt upon reaching the goal state.

\begin{figure}[htb!]
    \label{fig:P-set_learned_online}
    \centering
    \includegraphics[width=1.0\columnwidth]{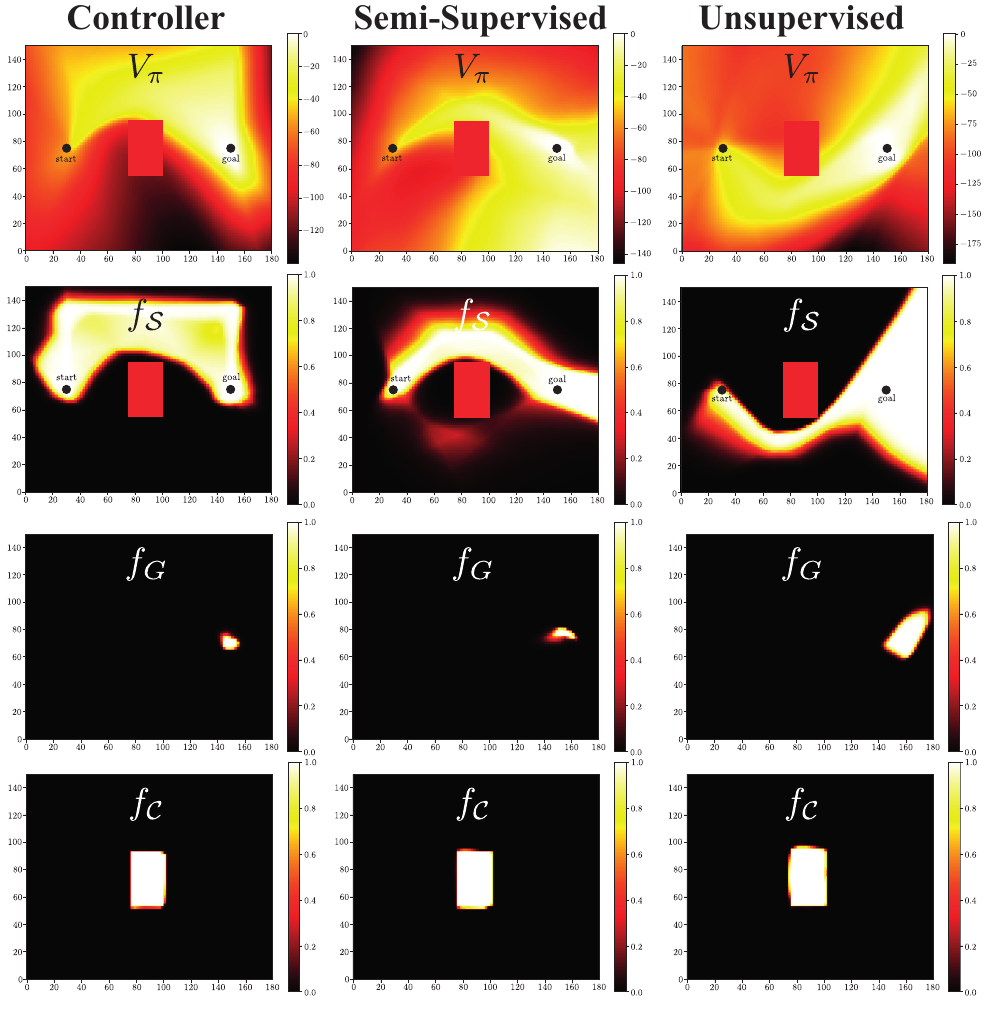}
    \caption{Heatmaps for each set of demonstrations after 50 online training episodes using algorithm \ref{alg:safe-set}.}
\end{figure}

By incorporating diverse training data, the unsupervised experiment controller is able to develop a more accurate representation of its environment, as shown in the heatmaps of Figure \ref{fig:P-set_learned_online}. After 50 online updates, these heatmaps reveal a larger, more expressive value-function and safe-set, highlighting the advantages of using unsupervised learning and a more diverse dataset for safe-RL. This also promotes safer exploration online as the agent possesses a more comprehensive understanding of potential failure modes. This emphasizes a key aspect of offline learning, where for a fixed number of demonstrations, if the dataset primarily comprises discrete controller demonstrations, the agent can easily learn an initial suboptimal policy, shown by the initially superior performance in Figure \ref{fig:F-online_safe_RL}. However, as the set of demonstrations is less diverse, the agent struggles with safely expanding its exploration due to a lack of diversity in the demonstration dataset.



A general design challenge for our unsupervised approach is selecting the number of skills to learn. More skills allow for greater discrimination between goal-reaching and constraint-violating behavior but take longer to train. Invariably this choice is task specific, with more complex tasks generally requiring more priors in order to accurately capture the different types of behaviors required to learn the safe-set and policy. We examined the performance of SMM using 16, and 64 priors, but found that it required excessive time to converge to a useful solution. This issue was previously identified by \citet{ICML40} with CIC. It would be valuable to investigate the impact of incorporating CIC's MI decomposition with the state marginal matching objective in SMM, thereby enabling the inclusion of a $p^{*}$ term. 


\section{Conclusions and Limitations}
\label{sec:conclusions}

In this work we investigated the importance of controlling the quantity and quality of data used to train safe-RL agents and how this impacts their ability to learn effective goal-reaching policies online. We discovered that using too little data in the initial training phase can lead to the safe-set becoming too restrictive, preventing online exploration and causing the agent to become stuck at the initial state. To address this we proposed optimistic-forgetting a modification to the safe-set LMPC algorithm that promotes online exploration. 

Through further investigation, we found data quality plays a crucial role for safe-RL, and increasing dataset diversity can help to improve the learned representation, facilitating greater exploration online. To generate diverse datasets we presented an unsupervised data-collection approach for safe-RL that enables the acquisition of complex behaviors without the need for hand-coded RL datasets. To achieve this we use competence-based unsupervised RL methods for both exploration and behavior classification.

A major limitation of our approach is its scalability to more complex tasks. We found that training the SMM agent is relatively computationally expensive compared to generating data from the controller alone. Additionally, as task complexity grows, more priors will likely be required to ensure the dataset provided to the safe-RL agent enables safe exploration online. Similar to the \textit{curse of dimensionality}, this problem has the potential to exponentially increase computational expense.

Additionally, practitioners are still required to make design decisions, such as: network design, state distribution $p^{*}$, entropy coefficients, skill representation $\mathcal{Z}$ and learning rate. All of which are important to be set appropriately to ensure the data collected balances task progress and exploration. Moving forward, we believe research should focus on creating algorithms that not only encourage exploration, but also offer improved representations of useful skills in order to make facilitate better use of unsupervised RL.

\bibliography{safe_rl}

\begin{thebibliography}{46}
\providecommand{\natexlab}[1]{#1}
\providecommand{\url}[1]{\texttt{#1}}
\expandafter\ifx\csname urlstyle\endcsname\relax
  \providecommand{\doi}[1]{doi: #1}\else
  \providecommand{\doi}{doi: \begingroup \urlstyle{rm}\Url}\fi

\bibitem[Bharadhwaj et~al.(2020)Bharadhwaj, Kumar, Rhinehart, Levine, Shkurti,
  and Garg]{ICML48}
Bharadhwaj, H., Kumar, A., Rhinehart, N., Levine, S., Shkurti, F., and Garg, A.
\newblock Conservative safety critics for exploration.
\newblock \emph{CoRR}, abs/2010.14497, 2020.
\newblock URL \url{https://arxiv.org/abs/2010.14497}.

\bibitem[Bishop(2007)]{ICML33}
Bishop, C.~M.
\newblock \emph{Pattern Recognition and Machine Learning (Information Science
  and Statistics)}.
\newblock Springer, 1 edition, 2007.
\newblock ISBN 0387310738.

\bibitem[Chua et~al.(2018)Chua, Calandra, McAllister, and Levine]{ICML38}
Chua, K., Calandra, R., McAllister, R., and Levine, S.
\newblock Deep reinforcement learning in a handful of trials using
  probabilistic dynamics models.
\newblock \emph{CoRR}, abs/1805.12114, 2018.
\newblock URL \url{http://arxiv.org/abs/1805.12114}.

\bibitem[Dalal et~al.(2018)Dalal, Dvijotham, Vecer{\'{\i}}k, Hester, Paduraru,
  and Tassa]{ICML47}
Dalal, G., Dvijotham, K., Vecer{\'{\i}}k, M., Hester, T., Paduraru, C., and
  Tassa, Y.
\newblock Safe exploration in continuous action spaces.
\newblock \emph{CoRR}, abs/1801.08757, 2018.
\newblock URL \url{http://arxiv.org/abs/1801.08757}.

\bibitem[Dasari et~al.(2019)Dasari, Ebert, Tian, Nair, Bucher, Schmeckpeper,
  Singh, Levine, and Finn]{ICML31}
Dasari, S., Ebert, F., Tian, S., Nair, S., Bucher, B., Schmeckpeper, K., Singh,
  S., Levine, S., and Finn, C.
\newblock Robonet: Large-scale multi-robot learning.
\newblock \emph{CoRR}, abs/1910.11215, 2019.
\newblock URL \url{http://arxiv.org/abs/1910.11215}.

\bibitem[Dulac{-}Arnold et~al.(2019)Dulac{-}Arnold, Mankowitz, and
  Hester]{ICML35}
Dulac{-}Arnold, G., Mankowitz, D.~J., and Hester, T.
\newblock Challenges of real-world reinforcement learning.
\newblock \emph{CoRR}, abs/1904.12901, 2019.
\newblock URL \url{http://arxiv.org/abs/1904.12901}.

\bibitem[Eysenbach et~al.(2018)Eysenbach, Gupta, Ibarz, and Levine]{ICML21}
Eysenbach, B., Gupta, A., Ibarz, J., and Levine, S.
\newblock Diversity is all you need: Learning skills without a reward function.
\newblock \emph{CoRR}, abs/1802.06070, 2018.
\newblock URL \url{http://arxiv.org/abs/1802.06070}.

\bibitem[Fu et~al.(2020)Fu, Kumar, Nachum, Tucker, and Levine]{ICML29}
Fu, J., Kumar, A., Nachum, O., Tucker, G., and Levine, S.
\newblock {D4RL:} datasets for deep data-driven reinforcement learning.
\newblock \emph{CoRR}, abs/2004.07219, 2020.
\newblock URL \url{https://arxiv.org/abs/2004.07219}.

\bibitem[Gros et~al.(2020)Gros, Zanon, and Bemporad]{ICML32}
Gros, S., Zanon, M., and Bemporad, A.
\newblock Safe reinforcement learning via projection on a safe set: How to
  achieve optimality?, 2020.
\newblock URL \url{https://arxiv.org/abs/2004.00915}.

\bibitem[G{\"{u}}l{\c{c}}ehre et~al.(2020)G{\"{u}}l{\c{c}}ehre, Wang, Novikov,
  Paine, Colmenarejo, Zolna, Agarwal, Merel, Mankowitz, Paduraru,
  Dulac{-}Arnold, Li, Norouzi, Hoffman, Nachum, Tucker, Heess, and
  de~Freitas]{ICML28}
G{\"{u}}l{\c{c}}ehre, {\c{C}}., Wang, Z., Novikov, A., Paine, T.~L.,
  Colmenarejo, S.~G., Zolna, K., Agarwal, R., Merel, J., Mankowitz, D.~J.,
  Paduraru, C., Dulac{-}Arnold, G., Li, J., Norouzi, M., Hoffman, M., Nachum,
  O., Tucker, G., Heess, N., and de~Freitas, N.
\newblock {RL} unplugged: Benchmarks for offline reinforcement learning.
\newblock \emph{CoRR}, abs/2006.13888, 2020.
\newblock URL \url{https://arxiv.org/abs/2006.13888}.

\bibitem[Gulcehre et~al.(2022)Gulcehre, Srinivasan, Sygnowski, Ostrovski,
  Farajtabar, Hoffman, Pascanu, and Doucet]{ICML36}
Gulcehre, C., Srinivasan, S., Sygnowski, J., Ostrovski, G., Farajtabar, M.,
  Hoffman, M., Pascanu, R., and Doucet, A.
\newblock An empirical study of implicit regularization in deep offline rl,
  2022.
\newblock URL \url{https://arxiv.org/abs/2207.02099}.

\bibitem[Haarnoja et~al.(2018{\natexlab{a}})Haarnoja, Zhou, Abbeel, and
  Levine]{ICML19}
Haarnoja, T., Zhou, A., Abbeel, P., and Levine, S.
\newblock Soft actor-critic: Off-policy maximum entropy deep reinforcement
  learning with a stochastic actor.
\newblock \emph{CoRR}, abs/1801.01290, 2018{\natexlab{a}}.
\newblock URL \url{http://arxiv.org/abs/1801.01290}.

\bibitem[Haarnoja et~al.(2018{\natexlab{b}})Haarnoja, Zhou, Abbeel, and
  Levine]{ICML42}
Haarnoja, T., Zhou, A., Abbeel, P., and Levine, S.
\newblock Soft actor-critic: Off-policy maximum entropy deep reinforcement
  learning with a stochastic actor.
\newblock \emph{CoRR}, abs/1801.01290, 2018{\natexlab{b}}.
\newblock URL \url{http://arxiv.org/abs/1801.01290}.

\bibitem[Haarnoja et~al.(2018{\natexlab{c}})Haarnoja, Zhou, Hartikainen,
  Tucker, Ha, Tan, Kumar, Zhu, Gupta, Abbeel, and Levine]{ICML2}
Haarnoja, T., Zhou, A., Hartikainen, K., Tucker, G., Ha, S., Tan, J., Kumar,
  V., Zhu, H., Gupta, A., Abbeel, P., and Levine, S.
\newblock Soft actor-critic algorithms and applications.
\newblock \emph{CoRR}, abs/1812.05905, 2018{\natexlab{c}}.
\newblock URL \url{http://arxiv.org/abs/1812.05905}.

\bibitem[Hansen et~al.(2019)Hansen, Dabney, Barreto, de~Wiele, Warde{-}Farley,
  and Mnih]{ICML45}
Hansen, S., Dabney, W., Barreto, A., de~Wiele, T.~V., Warde{-}Farley, D., and
  Mnih, V.
\newblock Fast task inference with variational intrinsic successor features.
\newblock \emph{CoRR}, abs/1906.05030, 2019.
\newblock URL \url{http://arxiv.org/abs/1906.05030}.

\bibitem[Higgins et~al.(2017)Higgins, Matthey, Pal, Burgess, Glorot, Botvinick,
  Mohamed, and Lerchner]{ICML39}
Higgins, I., Matthey, L., Pal, A., Burgess, C., Glorot, X., Botvinick, M.,
  Mohamed, S., and Lerchner, A.
\newblock beta-{VAE}: Learning basic visual concepts with a constrained
  variational framework.
\newblock In \emph{International Conference on Learning Representations}, 2017.
\newblock URL \url{https://openreview.net/forum?id=Sy2fzU9gl}.

\bibitem[Hoque et~al.(2021)Hoque, Balakrishna, Novoseller, Wilcox, Brown, and
  Goldberg]{ICML30}
Hoque, R., Balakrishna, A., Novoseller, E.~R., Wilcox, A., Brown, D.~S., and
  Goldberg, K.
\newblock Thriftydagger: Budget-aware novelty and risk gating for interactive
  imitation learning.
\newblock \emph{CoRR}, abs/2109.08273, 2021.
\newblock URL \url{https://arxiv.org/abs/2109.08273}.

\bibitem[Kalashnikov et~al.(2018)Kalashnikov, Irpan, Pastor, Ibarz, Herzog,
  Jang, Quillen, Holly, Kalakrishnan, Vanhoucke, and Levine]{ICML3}
Kalashnikov, D., Irpan, A., Pastor, P., Ibarz, J., Herzog, A., Jang, E.,
  Quillen, D., Holly, E., Kalakrishnan, M., Vanhoucke, V., and Levine, S.
\newblock Qt-opt: Scalable deep reinforcement learning for vision-based robotic
  manipulation.
\newblock \emph{CoRR}, abs/1806.10293, 2018.
\newblock URL \url{http://arxiv.org/abs/1806.10293}.

\bibitem[Lambert et~al.(2022)Lambert, Wulfmeier, Whitney, Byravan, Bloesch,
  Dasagi, Hertweck, and Riedmiller]{ICML25}
Lambert, N., Wulfmeier, M., Whitney, W.~F., Byravan, A., Bloesch, M., Dasagi,
  V., Hertweck, T., and Riedmiller, M.~A.
\newblock The challenges of exploration for offline reinforcement learning.
\newblock \emph{CoRR}, abs/2201.11861, 2022.
\newblock URL \url{https://arxiv.org/abs/2201.11861}.

\bibitem[Laskin et~al.(2021)Laskin, Yarats, Liu, Lee, Zhan, Lu, Cang, Pinto,
  and Abbeel]{ICML17}
Laskin, M., Yarats, D., Liu, H., Lee, K., Zhan, A., Lu, K., Cang, C., Pinto,
  L., and Abbeel, P.
\newblock {URLB}: Unsupervised reinforcement learning benchmark.
\newblock In \emph{Thirty-fifth Conference on Neural Information Processing
  Systems Datasets and Benchmarks Track (Round 2)}, 2021.
\newblock URL \url{https://openreview.net/forum?id=lwrPkQP_is}.

\bibitem[Laskin et~al.(2022)Laskin, Liu, Peng, Yarats, Rajeswaran, and
  Abbeel]{ICML40}
Laskin, M., Liu, H., Peng, X.~B., Yarats, D., Rajeswaran, A., and Abbeel, P.
\newblock {CIC:} contrastive intrinsic control for unsupervised skill
  discovery.
\newblock \emph{CoRR}, abs/2202.00161, 2022.
\newblock URL \url{https://arxiv.org/abs/2202.00161}.

\bibitem[Lee et~al.(2021)Lee, Devin, Zhou, Lampe, Bousmalis, Springenberg,
  Byravan, Abdolmaleki, Gileadi, Khosid, Fantacci, Chen, Raju, Jeong, Neunert,
  Laurens, Saliceti, Casarini, Riedmiller, Hadsell, and Nori]{ICML1}
Lee, A.~X., Devin, C., Zhou, Y., Lampe, T., Bousmalis, K., Springenberg, J.~T.,
  Byravan, A., Abdolmaleki, A., Gileadi, N., Khosid, D., Fantacci, C., Chen,
  J.~E., Raju, A., Jeong, R., Neunert, M., Laurens, A., Saliceti, S., Casarini,
  F., Riedmiller, M.~A., Hadsell, R., and Nori, F.
\newblock Beyond pick-and-place: Tackling robotic stacking of diverse shapes.
\newblock \emph{CoRR}, abs/2110.06192, 2021.
\newblock URL \url{https://arxiv.org/abs/2110.06192}.

\bibitem[Lee et~al.(2019)Lee, Eysenbach, Parisotto, Xing, Levine, and
  Salakhutdinov]{ICML22}
Lee, L., Eysenbach, B., Parisotto, E., Xing, E.~P., Levine, S., and
  Salakhutdinov, R.
\newblock Efficient exploration via state marginal matching.
\newblock \emph{CoRR}, abs/1906.05274, 2019.
\newblock URL \url{http://arxiv.org/abs/1906.05274}.

\bibitem[Levine(2021)]{ICML26}
Levine, S.
\newblock Understanding the world through action.
\newblock \emph{CoRR}, abs/2110.12543, 2021.
\newblock URL \url{https://arxiv.org/abs/2110.12543}.

\bibitem[Levine et~al.(2015)Levine, Finn, Darrell, and Abbeel]{ICML7}
Levine, S., Finn, C., Darrell, T., and Abbeel, P.
\newblock End-to-end training of deep visuomotor policies.
\newblock \emph{CoRR}, abs/1504.00702, 2015.
\newblock URL \url{http://arxiv.org/abs/1504.00702}.

\bibitem[Levine et~al.(2020)Levine, Kumar, Tucker, and Fu]{ICML34}
Levine, S., Kumar, A., Tucker, G., and Fu, J.
\newblock Offline reinforcement learning: Tutorial, review, and perspectives on
  open problems.
\newblock \emph{CoRR}, abs/2005.01643, 2020.
\newblock URL \url{https://arxiv.org/abs/2005.01643}.

\bibitem[Liu \& Abbeel(2021{\natexlab{a}})Liu and Abbeel]{ICML23}
Liu, H. and Abbeel, P.
\newblock {APS:} active pretraining with successor features.
\newblock \emph{CoRR}, abs/2108.13956, 2021{\natexlab{a}}.
\newblock URL \url{https://arxiv.org/abs/2108.13956}.

\bibitem[Liu \& Abbeel(2021{\natexlab{b}})Liu and Abbeel]{ICML44}
Liu, H. and Abbeel, P.
\newblock Behavior from the void: Unsupervised active pre-training.
\newblock \emph{CoRR}, abs/2103.04551, 2021{\natexlab{b}}.
\newblock URL \url{https://arxiv.org/abs/2103.04551}.

\bibitem[Mataric(1994)]{ICML5}
Mataric, M.~J.
\newblock Reward functions for accelerated learning.
\newblock In Cohen, W.~W. and Hirsh, H. (eds.), \emph{Machine Learning
  Proceedings 1994}, pp.\  181--189. Morgan Kaufmann, San Francisco (CA), 1994.
\newblock ISBN 978-1-55860-335-6.
\newblock \doi{https://doi.org/10.1016/B978-1-55860-335-6.50030-1}.
\newblock URL
  \url{https://www.sciencedirect.com/science/article/pii/B9781558603356500301}.

\bibitem[Nair et~al.(2018)Nair, Pong, Dalal, Bahl, Lin, and Levine]{ICML8}
Nair, A., Pong, V., Dalal, M., Bahl, S., Lin, S., and Levine, S.
\newblock Visual reinforcement learning with imagined goals.
\newblock \emph{CoRR}, abs/1807.04742, 2018.
\newblock URL \url{http://arxiv.org/abs/1807.04742}.

\bibitem[OpenAI et~al.(2018)OpenAI, Andrychowicz, Baker, Chociej,
  J{\'{o}}zefowicz, McGrew, Pachocki, Petron, Plappert, Powell, Ray, Schneider,
  Sidor, Tobin, Welinder, Weng, and Zaremba]{ICML4}
OpenAI, Andrychowicz, M., Baker, B., Chociej, M., J{\'{o}}zefowicz, R., McGrew,
  B., Pachocki, J., Petron, A., Plappert, M., Powell, G., Ray, A., Schneider,
  J., Sidor, S., Tobin, J., Welinder, P., Weng, L., and Zaremba, W.
\newblock Learning dexterous in-hand manipulation.
\newblock \emph{CoRR}, abs/1808.00177, 2018.
\newblock URL \url{http://arxiv.org/abs/1808.00177}.

\bibitem[Richards et~al.(2018)Richards, Berkenkamp, and Krause]{ICML10}
Richards, S.~M., Berkenkamp, F., and Krause, A.
\newblock The lyapunov neural network: Adaptive stability certification for
  safe learning of dynamic systems.
\newblock \emph{CoRR}, abs/1808.00924, 2018.
\newblock URL \url{http://arxiv.org/abs/1808.00924}.

\bibitem[Riedmiller et~al.(2021)Riedmiller, Springenberg, Hafner, and
  Heess]{ICML27}
Riedmiller, M.~A., Springenberg, J.~T., Hafner, R., and Heess, N.
\newblock Collect {\&} infer - a fresh look at data-efficient reinforcement
  learning.
\newblock \emph{CoRR}, abs/2108.10273, 2021.
\newblock URL \url{https://arxiv.org/abs/2108.10273}.

\bibitem[Rosolia \& Borrelli(2016)Rosolia and Borrelli]{ICML9}
Rosolia, U. and Borrelli, F.
\newblock Learning model predictive control for iterative tasks.
\newblock \emph{CoRR}, abs/1609.01387, 2016.
\newblock URL \url{http://arxiv.org/abs/1609.01387}.

\bibitem[Rosolia \& Borrelli(2019)Rosolia and Borrelli]{ICML15}
Rosolia, U. and Borrelli, F.
\newblock Sample-based learning model predictive control for linear uncertain
  systems.
\newblock \emph{CoRR}, abs/1904.06432, 2019.
\newblock URL \url{http://arxiv.org/abs/1904.06432}.

\bibitem[Rosolia et~al.(2018)Rosolia, Zhang, and Borrelli]{ICML14}
Rosolia, U., Zhang, X., and Borrelli, F.
\newblock A stochastic mpc approach with application to iterative learning.
\newblock In \emph{2018 IEEE Conference on Decision and Control (CDC)}, pp.\
  5152--5157, 12 2018.
\newblock \doi{10.1109/CDC.2018.8619268}.

\bibitem[Sharot(2011)]{ICML41}
Sharot, T.
\newblock The optimism bias.
\newblock \emph{Current Biology}, 21\penalty0 (23):\penalty0 R941--R945, 2011.
\newblock ISSN 0960-9822.
\newblock \doi{https://doi.org/10.1016/j.cub.2011.10.030}.
\newblock URL
  \url{https://www.sciencedirect.com/science/article/pii/S0960982211011912}.

\bibitem[Singh et~al.(2003)Singh, Misra, Hnizdo, Fedorowicz, and
  Demchuk]{ICML43}
Singh, H., Misra, N., Hnizdo, V., Fedorowicz, A., and Demchuk, E.
\newblock Nearest neighbor estimates of entropy.
\newblock \emph{American Journal of Mathematical and Management Sciences},
  23\penalty0 (3-4):\penalty0 301--321, 2003.
\newblock \doi{10.1080/01966324.2003.10737616}.
\newblock URL \url{https://doi.org/10.1080/01966324.2003.10737616}.

\bibitem[Tassa et~al.(2020)Tassa, Tunyasuvunakool, Muldal, Doron, Liu, Bohez,
  Merel, Erez, Lillicrap, and Heess]{ICML37}
Tassa, Y., Tunyasuvunakool, S., Muldal, A., Doron, Y., Liu, S., Bohez, S.,
  Merel, J., Erez, T., Lillicrap, T.~P., and Heess, N.
\newblock dm{\_}control: Software and tasks for continuous control.
\newblock \emph{CoRR}, abs/2006.12983, 2020.
\newblock URL \url{https://arxiv.org/abs/2006.12983}.

\bibitem[Thananjeyan et~al.(2019)Thananjeyan, Balakrishna, Rosolia, Li,
  McAllister, Gonzalez, Levine, Borrelli, and Goldberg]{ICML12}
Thananjeyan, B., Balakrishna, A., Rosolia, U., Li, F., McAllister, R.,
  Gonzalez, J.~E., Levine, S., Borrelli, F., and Goldberg, K.
\newblock Extending deep model predictive control with safety augmented value
  estimation from demonstrations (saved).
\newblock \emph{CoRR}, abs/1905.13402, 2019.
\newblock URL \url{http://arxiv.org/abs/1905.13402}.

\bibitem[Thananjeyan et~al.(2020)Thananjeyan, Balakrishna, Nair, Luo,
  Srinivasan, Hwang, Gonzalez, Ibarz, Finn, and Goldberg]{ICML16}
Thananjeyan, B., Balakrishna, A., Nair, S., Luo, M., Srinivasan, K., Hwang, M.,
  Gonzalez, J.~E., Ibarz, J., Finn, C., and Goldberg, K.
\newblock Recovery {RL:} safe reinforcement learning with learned recovery
  zones.
\newblock \emph{CoRR}, abs/2010.15920, 2020.
\newblock URL \url{https://arxiv.org/abs/2010.15920}.

\bibitem[Tian et~al.(2020)Tian, Nair, Ebert, Dasari, Eysenbach, Finn, and
  Levine]{ICML6}
Tian, S., Nair, S., Ebert, F., Dasari, S., Eysenbach, B., Finn, C., and Levine,
  S.
\newblock Model-based visual planning with self-supervised functional
  distances.
\newblock \emph{CoRR}, abs/2012.15373, 2020.
\newblock URL \url{https://arxiv.org/abs/2012.15373}.

\bibitem[van~den Oord et~al.(2018)van~den Oord, Li, and Vinyals]{ICML46}
van~den Oord, A., Li, Y., and Vinyals, O.
\newblock Representation learning with contrastive predictive coding.
\newblock \emph{CoRR}, abs/1807.03748, 2018.
\newblock URL \url{http://arxiv.org/abs/1807.03748}.

\bibitem[Wilcox et~al.(2021)Wilcox, Balakrishna, Thananjeyan, Gonzalez, and
  Goldberg]{ICML11}
Wilcox, A., Balakrishna, A., Thananjeyan, B., Gonzalez, J.~E., and Goldberg, K.
\newblock {LS3:} latent space safe sets for long-horizon visuomotor control of
  iterative tasks.
\newblock \emph{CoRR}, abs/2107.04775, 2021.
\newblock URL \url{https://arxiv.org/abs/2107.04775}.

\bibitem[Yarats et~al.(2021)Yarats, Fergus, Lazaric, and Pinto]{ICML20}
Yarats, D., Fergus, R., Lazaric, A., and Pinto, L.
\newblock Mastering visual continuous control: Improved data-augmented
  reinforcement learning.
\newblock \emph{CoRR}, abs/2107.09645, 2021.
\newblock URL \url{https://arxiv.org/abs/2107.09645}.

\bibitem[Yarats et~al.(2022)Yarats, Brandfonbrener, Liu, Laskin, Abbeel,
  Lazaric, and Pinto]{ICML24}
Yarats, D., Brandfonbrener, D., Liu, H., Laskin, M., Abbeel, P., Lazaric, A.,
  and Pinto, L.
\newblock Don't change the algorithm, change the data: Exploratory data for
  offline reinforcement learning.
\newblock \emph{CoRR}, abs/2201.13425, 2022.
\newblock URL \url{https://arxiv.org/abs/2201.13425}.

\end{thebibliography}
\bibliographystyle{icml2022}

\newpage
\appendix
\onecolumn
\section{Test Environments and Controllers}
\label{sec:te_and_c}

All test environments focus on navigating to a goal-reaching state around a central block constraint. Tasks are made increasingly difficult by increasing the complexity of the agents action and observation space.

\subsection{Simple Point Bot (SPB)}

The Simple Point Bot (SPB) task requires the agent to navigate to a goal state without colliding with the constraint. The agent's state is fully observable and controlled directly through its velocity.

\subsubsection{Environment Definition}
\begin{itemize}
    \item \textbf{State}: $[x, y]$
    \item \textbf{Action}: $[u, v]$
    \item \textbf{Goal}: $[x_{g}, y_{g}]$
\end{itemize}

\begin{figure}[htb!]
    \label{fig:spb}
    \centering
    \includegraphics[width=0.5\textwidth]{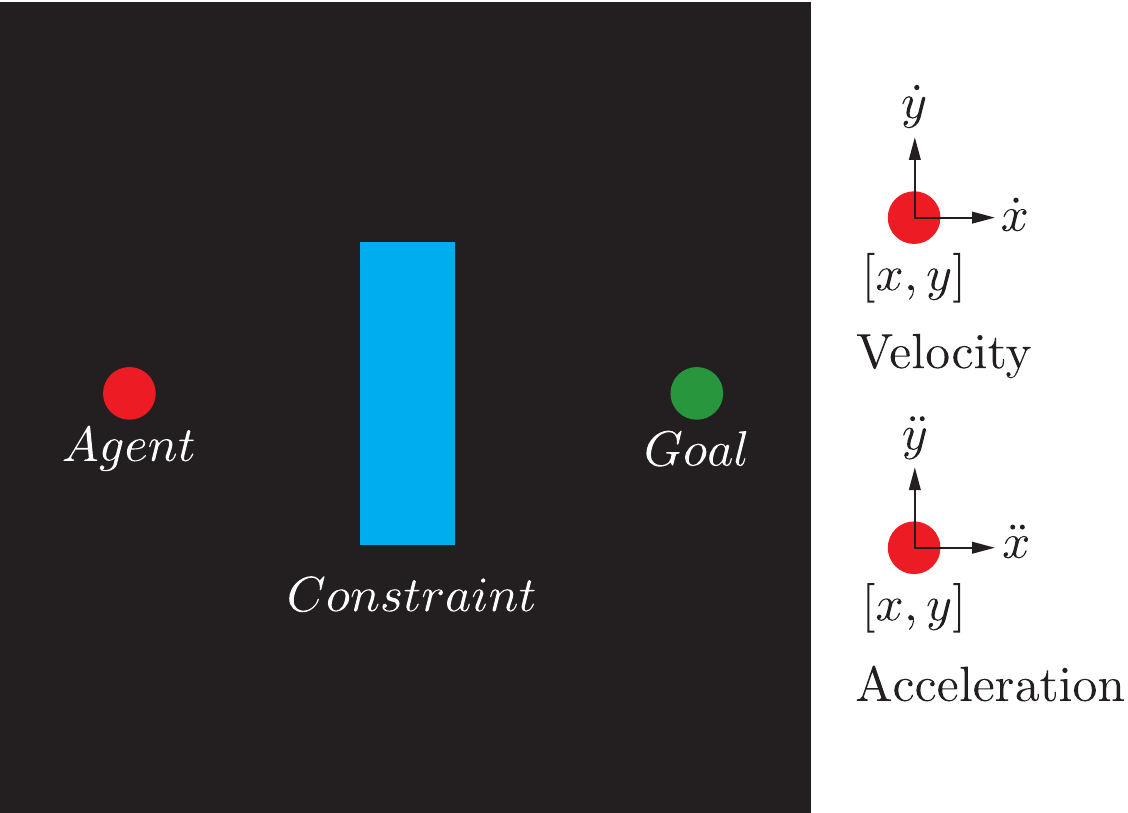}
    \caption{PointBot environments}
\end{figure}

\begin{figure}[htb!]
    \label{fig:bottleneck}
    \centering
    \includegraphics[width=0.5\textwidth]{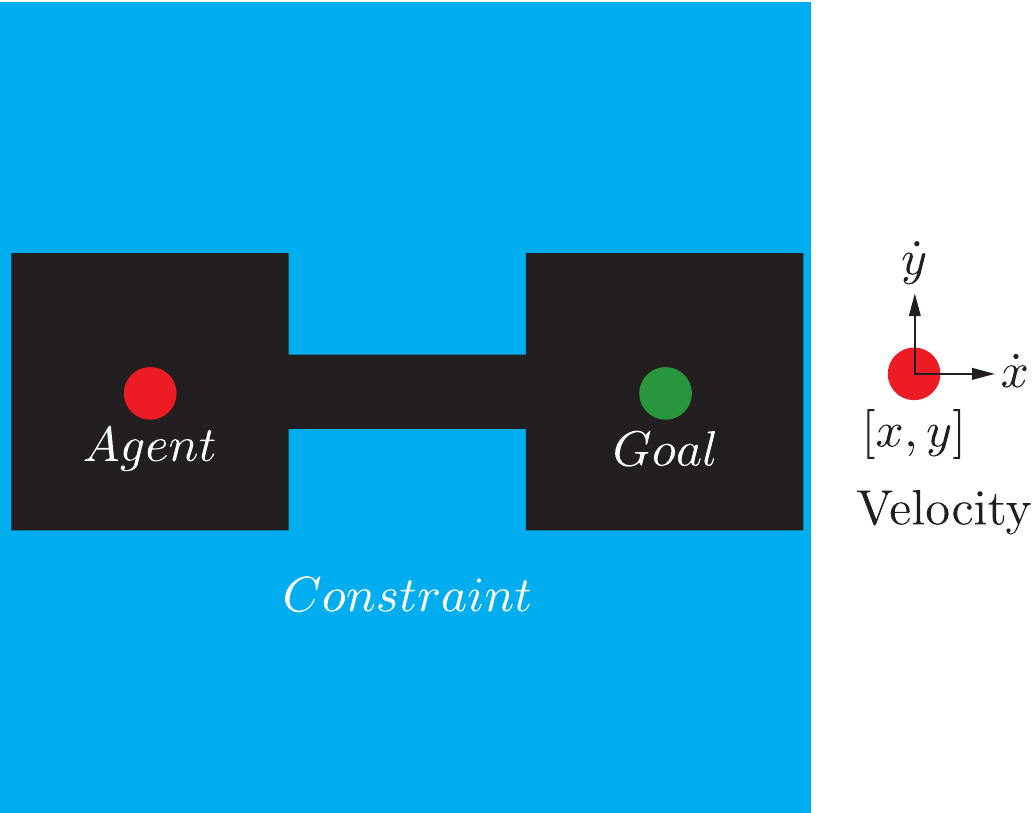}
    \caption{Bottleneck environment}
\end{figure}

\subsubsection{Hand-Controllers}

\begin{figure}[htb!]
    \label{fig:spb_control}
    \centering
    \includegraphics[width=0.2\textwidth]{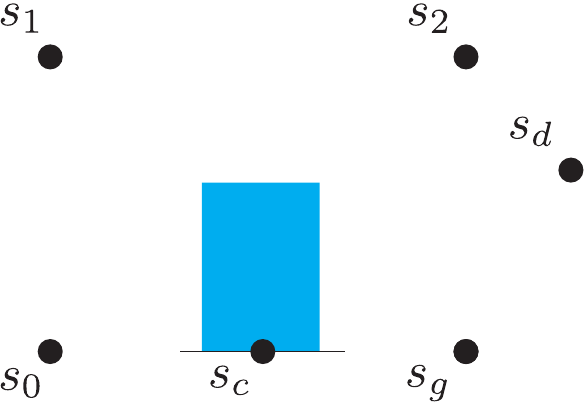}
    \caption{Controller defined points}
\end{figure}

Algorithms \ref{alg:gr-spb} and \ref{alg:cv-spb} are used to generate CV and GR samples based on the keypoints outlined in figure \ref{fig:spb_control}.

\begin{algorithm}[htb!]
    \caption{SPBController - Goal Reaching}
    \label{alg:gr-spb}
    \begin{algorithmic}
        \REQUIRE{goal position $s_{g}$, intermediary positions $s_{1}$, $s_{2}$, max action size $\mathbf{a}_{max}$}
        \INPUT{time $t$, state $s_{t}$}
        \IF{$t < 20$}
            \STATE $a_{t} = \mathbf{a}_{max}|s_{1} - s_{t}|$
            \STATE $t \mathrel{+}= 1$
        \ENDIF
        \IF{$21 < t < 60$}
            \STATE $a_{t} = \mathbf{a}_{max}|s_{2} - s_{t}|$
            \STATE $t \mathrel{+}= 1$
        \ENDIF
        \IF{$t > 60$}
            \STATE $a_{t} = \mathbf{a}_{max}|s_{g} - s_{t}|$
        \ENDIF
        \OUTPUT{action $a_{t}$}
    \end{algorithmic}
\end{algorithm}

\begin{algorithm}[htb!]
    \caption{SPBController - Constraint Violating}
    \label{alg:cv-spb}
    \begin{algorithmic}
        \REQUIRE{constraint center $s_{c}$, random position $s_{d}$}
        \INPUT{time $t$, state $s_{t}$}
        \IF{$t < 15$}
            \STATE $a_{t} = \mathbf{a}_{max}|s_{d} - s_{t}|$
            \STATE $t \mathrel{+}= 1$
        \ENDIF
        \IF{$t > 15$}
            \STATE $a_{t} = \mathbf{a}_{max}|s_{c} - s_{t}|$
            \STATE $t \mathrel{+}= 1$
        \ENDIF
        \OUTPUT{action $a_{t}$}
    \end{algorithmic}
\end{algorithm}

\newpage

\subsection{Simple Velocity Bot (SVB)}

The Simple Velocity Bot (SVB) task uses the acceleration of the agent to control the state, and velocity is unobserved, therefore it is a partially observable environment.

\subsubsection{Environment Definition}
\begin{itemize}
    \item \textbf{State}: $[x, y]$
    \item \textbf{Action}: $[a_{x}, a_{y}]$
    \item \textbf{Goal}: $[x_{g}, y_{g}]$
\end{itemize}

\subsubsection{Hand-Controllers}

For the SVB controller we use a PID controller with gains scheduled based upon the agents position. With controller gains:

\begin{table}[htb!]
    \centering
    \caption{Controller Gains}
    \label{tab:A-gains}
    \begin{tabular}{l|lll}
             & \multicolumn{1}{l}{$K_{p}$} & $K_{i}$   & $K_{d}$  \\ \hline
    $s_{0}$ $\to$ $s_{2}$ & 5.0                     & 0.05 & 0.0 \\
    $s_{2}$ $\to$ $s_{g}$ & 5.0                     & 0.5  & 4.0
    \end{tabular}
\end{table}

The controller acts upon the velocity error $e_{t} = v_{t} - v_{dem}$, these are the same for $x$ and $y$ velocity components.

\begin{equation}
    \ddot{x}_{t} = K_{p} e_{t} + K_{i} \int_{0}^{t}e_{\tau} d\tau + K_{d} \dot{e}_{t}
\end{equation}

To navigate the agent around the environment to the goal state we then use a set of states $\mathbf{G} = (\mathbf{s}_{0}, \mathbf{s}_{1}, \mathbf{s}_{2}, \mathbf{s}_{g})$ and define two control modes whilst en-route to the final goal state. When the goal state is the next ensuing point the agent then just moves to the goal state. We define 3 points as part of this trajectory, based on the goal index $i_{g}$. We then work out a track angle to go from $a \to b$ and $b \to c$:

\begin{enumerate}
    \item Calculate track points: $\mathbf{s}_{a}=\mathbf{G}[i_{g}]$, $\mathbf{s}_{b}=\mathbf{G}[i_{g}+1]$ and $\mathbf{s}_{c}=\mathbf{G}[i_{g}+2]$
    \item Calculate track angle in and out: $\theta_{in} = \arccos(\frac{s_{a}}{|s_{a}|} \cdot \frac{\mathbf{s}_{b}}{|\mathbf{s}_{b}|})$, $\theta_{out} = \arccos(\frac{s_{b}}{|\mathbf{s}_{b}|} \cdot \frac{\mathbf{s}_{c}}{|\mathbf{s}_{c}|})$
    \item Calculate turn angle $\phi = \theta_{out} - \theta_{in}$
\end{enumerate}

\begin{figure}[htb!]
    \centering
    \label{fig:H-svb_turn}
    \includegraphics[width=0.3\textwidth]{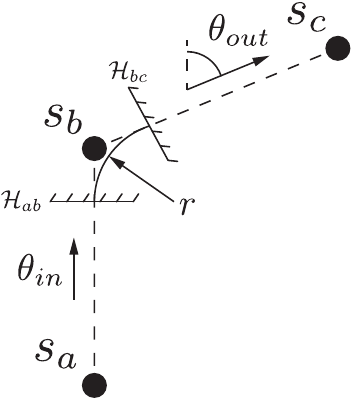}
    \caption{Navigation problem setup}
\end{figure}

Figure \ref{fig:H-svb_turn} labels the key points and desired track from $a \to b \to c$. We define 2 navigation modes, \textit{track-to-point} and \textit{fly-by-point}. In track-to-point the agent tracks a straight line to the target position, whilst in fly-by-point the agent aims to maintain a constant turn radius around 2 points bounded by planes based on the turn radius $r$. When the agent passes through a plane it changes mode.

\begin{figure}[htb!]
    \centering
    \label{fig:I-svb_track}
    \includegraphics[width=0.3\textwidth]{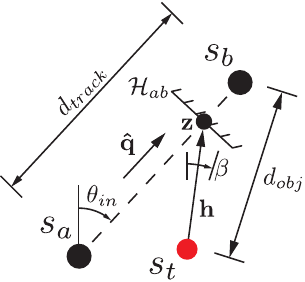}
    \caption{Track-to-Point key values and dimensions}
\end{figure}

Figure \ref{fig:I-svb_track} illustrates the key values and dimensions used by the track-to-point controller. If in \textit{track-to-point} mode, used from $\mathbf{s}_{a} \to \mathcal{H}_{ab}$:

\begin{enumerate}
    \item Define $\mathbf{\hat{q}} = \frac{\mathbf{s}_{b} - \mathbf{s}_{a}}{|\mathbf{s}_{b} - \mathbf{s}_{a}|}$ as the unit vector perpendicular to the plane $\mathcal{H}_{ab}$. 
    \item Define the point where the vector $\mathbf{s}_{b} - \mathbf{s}_{a}$ intersects the plane $\mathcal{H}_{ab}$ as: $\mathbf{z} = \mathbf{s}_{b} - \frac{r}{tan(\frac{\phi}{2})} \mathbf{\hat{q}}$.
    \item Calculate vector from agent to plane center: $\mathbf{h} = \mathbf{z} - \mathbf{s}_{t}$.
    \item Calculate which side of the plane the agent lies on: $h = \mathbf{h} \cdot \mathbf{\hat{q}}$.
    \item If $h > 0$, change to \textit{fly-by-point} mode on next iteration.
\end{enumerate}

To then calculate the heading allowing us to track the line from $\mathbf{s}_{a} \to \mathbf{z}$:

\begin{enumerate}
    \item Define angle from agent to $\mathbf{s}_{b}$ as $\beta = \arccos(\frac{\mathbf{s}_{t}}{|\mathbf{s}_{t}|} \cdot \frac{\mathbf{s}_{b}}{\mathbf{s}_{b}})$
    \item Define distance to the objective as: $d_{obj} = || \mathbf{s}_{t} - \mathbf{s}_{b} ||$
    \item Define track distance as: $d_{track} = ||\mathbf{s}_{b} - \mathbf{s}_{a}||$
    \item Calculate off track error as: $\phi = \beta - \theta_{in}$
    \item Calculate demanded heading as: $\theta_{hdg} = \frac{d_{track} \phi}{2 d_{obj}} + \theta_{in}$
\end{enumerate}

\begin{figure}[htb!]
    \centering
    \label{fig:J-svb_flyby}
    \includegraphics[width=0.6\textwidth]{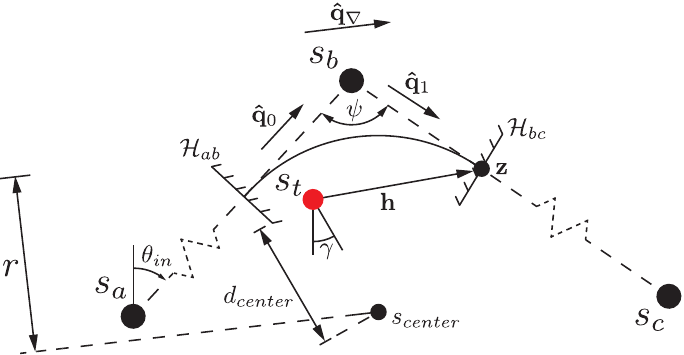}
    \caption{Fly-by-Point key values and dimensions}
\end{figure}

Figure \ref{fig:J-svb_flyby} illustrates the key values and dimensions used by the fly-by-point controller. If in \textit{fly-by-point} mode, used from $\mathcal{H}_{ab} \to \mathcal{H}_{bc}$:

\begin{enumerate}
    \item Define $\mathbf{\hat{q}}_{0} = \frac{\mathbf{s}_{b} - \mathbf{s}_{a}}{|\mathbf{s}_{b} - \mathbf{s}_{a}|}$ and $\mathbf{\hat{q}}_{1} = \frac{\mathbf{s}_{c} - \mathbf{s}_{b}}{|\mathbf{s}_{c} - \mathbf{s}_{b}|}$ as unit vectors perpendicular to $\mathcal{H}_{ab}$ and $\mathcal{H}_{bc}$ respectively. 
    \item Define $\mathbf{\hat{q}}_{\nabla} = \frac{\mathbf{\hat{q}}_{1} - \mathbf{\hat{q}}_{0}}{|\mathbf{\hat{q}}_{1} - \mathbf{\hat{q}}_{0}|}$ as the unit vector between $\mathcal{H}_{ab} \to \mathcal{H}_{bc}$.
    \item Calculate center point of turn fillet as: $\mathbf{s}_{center} = \mathbf{s}_{b} - \frac{r}{\sin(\frac{\psi}{2})} \mathbf{\hat{q}}_{\nabla}$.
    \item Define point where the vector $\mathbf{s}_{c} - \mathbf{s}_{a}$ intersects the plane $\mathcal{H}_{bc}$ as: $\mathbf{z} = \mathbf{s}_{b} + \frac{r}{\tan(\frac{\psi}{2})} \mathbf{\hat{q}}_{1}$.
    \item Calculate vector from agent to plane center: $\mathbf{h} = \mathbf{z} - \mathbf{s}_{t}$.
    \item Calculate which side of the plane the agent lies on: $h = \mathbf{h} \cdot \mathbf{\hat{q}}_{1}$.
    \item If $h > 0$, change to \textit{track-to-point} mode on next iteration.
\end{enumerate}

To then calculate the heading, allowing us to track along the arc from $\mathcal{H}_{ab} \to \mathbf{z}$ with radius $r$:

\begin{enumerate}
    \item Define $d_{center} = ||\mathbf{s}_{t} - \mathbf{s}_{center}||$ as the distance of the agent from the center of the fillet.
    \item Define circle angle as: $\gamma = \arctan(\mathbf{s}_{t} - \mathbf{s}_{center})$.
    \item Calculate circle error as: $e_{t} = \frac{\mathbf{s}_{center} - r}{r}$.
    \item Declare turning orbit gain: $k_{orbit} = 4.0$.
    \item Calculate heading as: $\theta_{heading} = \gamma + \frac{\pi}{2} - \arctan(k_{orbit} e_{t})$.
\end{enumerate}

If in the final state from $s_{2} \to s_{g}$ we simply drive the agent to the goal state and stop when close enough:

\begin{enumerate}
    \item Define heading and distance as: $d_{goal} = ||\mathbf{s}_{t} - \mathbf{s}_{g}||$, $\theta_{heading} = \arctan(\mathbf{s}_{t} - \mathbf{s}_{g})$.
    \item If $d_{goal} > 20.0$ then demanded speed is $|u|_{dem} = 0.075 d_{goal}^{2}$.
    \item Else If $d_{goal} < 10.0$ then demanded speed is $|u|_{dem} = 0.0$.
    \item Else demanded speed is $|u|_{dem} = 3.0$.
\end{enumerate}

To generate an action to feed to the agent we calculate the relative demanded velocity components based on $\theta_{heading}$ and use the PID gains from table \ref{tab:A-gains} to get values for $[\ddot{x}, \ddot{y}]$. To gather constraint-violating behavior samples we use the same approach as defined in algorithm \ref{alg:cv-spb} and use a PID controller to control the agent (using gains from the first row of table \ref{tab:A-gains}).

\section{Safe RL}
\label{sec:safeRL}


Implementation details for each component of the safe RL algorithm are described in the following section. 

\subsubsection{Variational AutoEncoder (VAE)}

For image based observations inputs are scaled to $(64, 64, 3)$ input images before being fed into the $\beta$-VAE, using a Convolutional Neural Network (CNN) for $f_{enc}$ and a transpose CNN for $f_{dec}$. Training is with mean-squared error (MSE) loss and Kullback-Leibler (KL) divergence, for $s_{t} \in \mathcal{S}$ the loss is:

\begin{equation}
    \label{eq:vae-cost}
    J(\theta) = \| f_{dec}(z_{t}) - s_{t} \|_{2}^{2} + \beta D_{KL} (f_{enc}(z_{t}|s_{t}) \| \mathcal{N}(0, 1))
\end{equation}

\subsubsection{Probabilistic Dynamics}

Based on PETS \citep{ICML38}, we trained a probabilistic ensemble of neural networks to learn the agent's dynamics. Trained with a max log-likelihood objective, each network has 2 hidden layers and 128 hidden units. The loss for states $s_{t}, s_{t+1} \in \mathcal{S}$ with action $a_{t} \in \mathcal{A}$ for dynamics model $f_{dyn, \theta}$ is:

\begin{equation}
    \label{eq:pets-cost}
    J(\theta) = -\log (f_{dyn, \theta} (s_{t+1}|s_{t}, a_{t}))
\end{equation}

We use the TS-1 method for planning with $f_{dyn}$.

\subsubsection{Value Functions}

We trained an ensemble of value functions using fully connected neural networks (consisting of 3 hidden layers and 256 hidden units) recursively to predict long-term reward. During offline training the value function is trained to predict the cost-to-go on all trajectories in $\mathcal{D}$. The training loss with the $\theta$ parameterized value function $V$ is given as:

\begin{equation}
    \label{eq:offline_value_fcn}
    J(\theta) = \biggl( V_{\theta}^{\pi}(s_{t}) - \sum_{i=1}^{T-t} \gamma^{i}r_{t+i} \biggr)^{2}
\end{equation}

In online training the target network $V^{\pi^{'}}$ is stored and the temporal difference (TD-1) error calculated:

\begin{equation}
    \label{eq:online_value_fcn}
    J(\theta) = (V_{\theta}^{\pi}(s_{t}) - (r_{t} + \gamma V_{\theta^{'}}^{\pi^{'}}(s_{t+1})))
\end{equation}

The value function is updated only using data from suboptimal demonstrations, or online data. In equation \ref{eq:online_value_fcn} $\theta^{'}$ is the parameters of a lagged target network and $\pi^{'}$ is the policy at the time-step $\theta^{'}$ was set, $\gamma = 0.99$ is the discount factor.

\subsubsection{Constraint and Goal Estimators}

The constraint indicator $f_{\mathcal{C}} : \mathcal{Z} \to \{ 0, 1 \}$ is represented by a neural network with 3 hidden layers and 256 hidden units. Each layer is trained with a binary cross entropy loss using demonstrations from $\mathcal{D}_{\text{constraint}}$ as unsafe examples and $\mathcal{D}_{\text{constraint}}^{'}$ as safe examples. 

The goal estimator $f_{\mathcal{G}} : \mathcal{Z} \to \{ 0, 1 \}$ similarly is represented by a neural network consisting of 3 hidden layers and 256 hidden units. Goal-reaching demonstrations are used to train the classifier from $\mathcal{D}_{gr}$. Both estimators use a binary cross entropy loss function for training. 

\subsubsection{Safe Set}

The safe set classifier $f_{\mathbb{S}}(s_{t})$ is represented by a neural network with 3 hidden layers and 256 hidden units, trained to predict:

$$
    f_{\mathbb{S}}(s_{t}) = \max (\mathds{1}_{\mathbb{S}}(s_{t}), \gamma_{\mathbb{S}} f_{\mathbb{S}}(s_{t+1}))
$$

Where $\mathds{1}(s_{t})$ is an indicator function, indicating whether $s_{t}$ is part of a successful trajectory. Training data is sampled uniformly from the entire dataset, $\mathcal{D}$.

\subsubsection{Model Based Planning}

The Cross Entropy Method (CEM) is used to solve the following optimization problem in order to select actions $a_{t:t+H-1}$:

\begin{equation}
    \label{eq:planning_optimization}
    \begin{aligned}
        \argmax_{a_{t:t+H-1}} \quad & \mathbb{E}_{s_{t:t+H}} \Biggl[ \sum_{i=1}^{H-1} f_{\mathcal{G}}(s_{t+i}) + V^{\pi}(s_{t+H}) \Biggr] \\
        \text{s.t.} \quad & s_{t+1} \sim f_{dyn}(s_{t+1}|s_{t}, a_{t}) \forall \in \{ t, \ldots, t+H-1 \} \\
        \quad & \hat{\mathbb{P}} \bigl( s_{t+H} \in \mathbb{S}_{\mathcal{S}} \bigr) \geq 1-\delta_{\mathbb{S}} \\
        \quad & \hat{\mathbb{P}} \bigl( s_{t+i} \in \mathbb{S}_{\mathcal{C}} \bigr) \leq \delta_{\mathcal{C}} \forall i \in \{ 0, \ldots, H-1 \}
    \end{aligned}
\end{equation}

Details of the application of the CEM method for model-based planning are provided in \citet{ICML38} and \citet{ICML11}. For image based observations equation \ref{eq:planning_optimization} is re-parameterized based on the VAE latent variable $\mathcal{Z}$, such that $f_{enc}(s_{t}) \to z_{t}$. Parameters used for the safe-RL procedure are provided in table \ref{tab:safeRL}.

\begin{table}[htb!]
    \centering
    \caption{Safe RL (LS$^{3}$) parameters}
    \label{tab:safeRL}
    \begin{tabular}{ll}
    \hline
    \textbf{SafeRL Parameter} & \textbf{Value}   \\ \hline
    SS Violation ($\delta_{\mathbb{S}}$)   & 0.8     \\
    Constraint Violation ($\delta_{\mathcal{C}}$) & 0.2     \\
    VAE Beta ($\beta$)             & $10^{-6}$ \\
    Planning Horizon ($H$)  & 5      \\
    CEM ($n_{\text{particle}}$)  & 20      \\
    CEM ($n_{\text{candidate}}$)      & 1000    \\
    CEM ($n_{\text{elite}}$) & 100     \\
    CEM iterations ($N_{\text{cem}}$)  & 5       \\
    CEM ($d$)                     & 32      \\
    CEM ($p_{\text{random}}$)              & 1.0     \\
    Frame Stacking        & No      \\
    Batch Size             & 0.01    \\
    Discount Factor ($\gamma$) & 0.99    \\
    SS discount ($\gamma_{\mathbb{S}}$)    & 0.3 \\
    \hline
    \end{tabular}
\end{table}

\section{Unsupervised RL}
\label{sec:unsupervised_comparison}

To generate a group of constraint violating and goal-reaching skills we made use of competence based unsupervised RL techniques. To learn this skill embeding we used Soft Actor Critic (SAC) \cite{ICML42}, an off-policy actor-critic deep RL algorithm. SAC augments the standard RL expected discounted sum of rewards objective $\sum_{t} \mathbb{E}_{(\mathbf{s}_{t}, \mathbf{a}_{t}) \sim \rho_{\pi}} [r(\mathbf{s}_{t}, \mathbf{a}_{t})]$ with the expected entropy over the policy $\rho_{\pi}(s_{t})$. This makes the policy objective:

\begin{equation}
    J(\pi) = \sum^{T}_{t=0} \mathbb{E}_{(\mathbf{s}_{t}, \mathbf{a}_{t}) \sim \rho_{\pi}} [r(\mathbf{s}_{t}, \mathbf{a}_{t}) + \alpha \mathcal{H}(\pi(\cdot | \mathbf{s}_{t})) ]
    \label{eq:sac-reward_again}
\end{equation}

Where $\alpha$ represents the importance of the entropy term against the CMDP reward term. All the unsupervised learning algorithms used in this work include an entropy maximization term $\mathcal{H}(\pi(\cdot | \mathbf{s}_{t}))$.

In order to find the optimal policy $\pi^{*}$ defined by the objective function in Equation \ref{eq:sac-reward_again} SAC uses a parameterized: state value function $V_{\psi}(\mathbf{s}_{t})$, soft Q-function $Q_{\theta}(\mathbf{s}_{t}, \mathbf{a}_{t})$ and tractable policy $\pi_{\phi}(\mathbf{a}_{t}|\mathbf{s}_{t})$. The soft value function is trained to minimize the residual square error:

$$
    J_{V}(\psi) = \mathbb{E}_{\mathbf{s}_{t} \sim \mathcal{D}} \Biggl[ \frac{1}{2} \Bigl( V_{\psi}(\mathbf{s}_{t}) - \mathbb{E}_{\mathbf{a}_{t} \sim \pi_{\phi}}[Q_{\theta}(\mathbf{s}_{t}, \mathbf{a}_{t}) - \log \pi_{\phi}(\mathbf{a}_{t}|\mathbf{s}_{t})]^{2} \Bigr) \Biggr]
$$

The soft Q-function is trained to minimize the soft Bellman error:

$$
    J_{Q}(\theta) = \mathbb{E}_{(\mathbf{s}_{t}, \mathbf{a}_{t}) \sim \mathcal{D}} \biggl[ \frac{1}{2} \bigl( Q_{\theta}(\mathbf{s}_{t}, \mathbf{a}_{t}) - \hat{Q}(\mathbf{s}_{t}, \mathbf{a}_{t}) \bigr) ^{2} \biggr]
$$

where,

$$
    \hat{Q}(\mathbf{s}_{t}, \mathbf{a}_{t}) = r(\mathbf{s}_{t}, \mathbf{a}_{t}) + \gamma \mathbb{E}_{\mathbf{s}_{t+1} \sim \mathcal{D}} \bigl[ V_{\overline{\psi}}(\mathbf{s}_{t+1}) \bigr]
$$

Bringing these together and using the reparameterization trick to represent actions with samples from a spherical Gaussian $\mathbf{a}_{t} = f_{\phi}(\epsilon_{t}; \mathbf{s}_{t})$ the policy objective becomes:

$$
    J_{\pi}(\phi) = \mathbb{E}_{\mathbf{s}_{t} \sim \mathcal{D}, \epsilon_{t} \sim \mathcal{N}} [ \log \pi_{\phi} (f_{\phi}(\epsilon_{t}; \mathbf{s}_{t})|\mathbf{s}_{t}) - Q_{\theta}(\mathbf{s}_{t}, f_{\phi}(\epsilon_{t}; \mathbf{s}_{t})) ]
$$

SAC then uses two Q-functions to reduce positive bias during policy improvement and alternates between adding experience into the replay buffer, with the policy $\pi_{\phi}(\mathbf{a}_{t}|\mathbf{s}_{t})$ and updating the policy via gradient descent. The hyperparameters used for the SAC implementation in this work are outlined below in Table \ref{tab:C-sac}. 

Compared to SAC, DrQ-V2 is more robust to reward scaling as exploration is not directly tied to the objective function but is instead controlled by applying Gaussian noise to the objective function. We found selecting an appropriate noise schedule difficult for the navigation tasks in this work because the agent needs many exploratory samples to understand how to reach the goal but relatively fine control to remain at the goal once it reached it. The exploratory noise values used in DrQ-v2 made it difficult for the agent to stay at the goal after it had reached it, but any lower noise values caused the agent to run into the block too often.

\begin{table}[htb!]
    \centering
    \label{tab:C-sac}
    \caption{Soft Actor Critic (SAC) Parameters}
    \label{Tab:sac-params}
    \begin{tabular}{ll}
    \hline
    \textbf{SAC Parameter}            & \textbf{Value} \\ \hline
    Replay Buffer Capacity            & $10^{6}$            \\
    Batch Size                        & 256            \\
    Discount ($\gamma$)               & 0.99           \\
    Learning Rate ($\lambda$)         & $10^{-3}$           \\
    Activation Function               & ReLU           \\
    Optimizer                         & Adam           \\
    Hidden Layers                     & 5              \\
    Hidden Dims                       & 64             \\
    Polyack Average ($\tau$)          & 0.005          \\
    Entropy Regularization ($\alpha$) & 0.1            \\
    Target Step Interval              & 1              \\
    Step Update Frequency             & 1              \\ \hline
    \end{tabular}
\end{table}

\begin{figure}[htb!]
    \centering
    \label{fig:L-SAC_comparison}
    \includegraphics[width=0.8\textwidth]{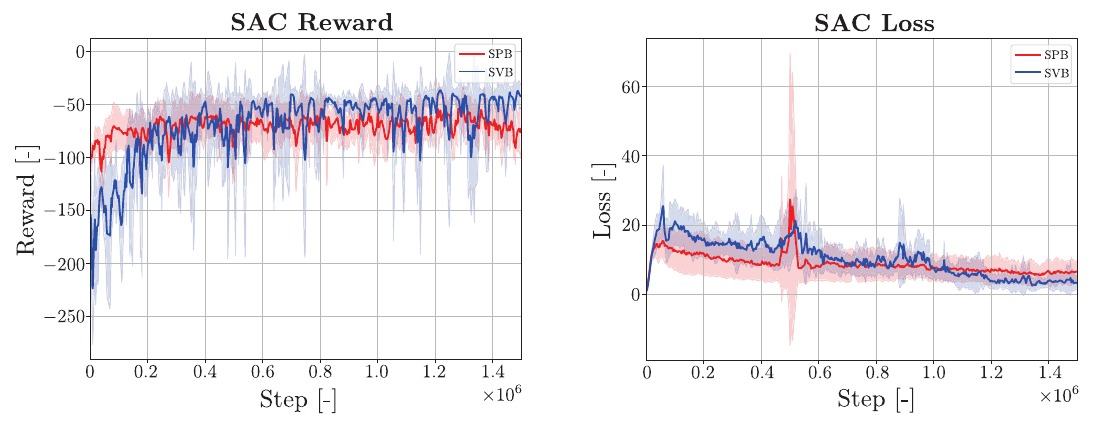}
    \caption{Comparison of SAC return and Q-loss using $L_{2}$ goal norm from equation \ref{eq:pstar} on SPB and SVB. SPB is deceptively difficult as the agent needs to explore outside the box to reach the maximum reward which moves it further from the goal resulting in temporarily lower return. Runs are taken over 5 random seeds and error bounds at 95\% confidence interval shown.}
\end{figure}

Figure \ref{fig:L-SAC_comparison} compares the training performance on SPB and SVB, we found that even though we can learn an effective goal-reaching policy for SVB with SAC. We could not learn useful goal reaching behavior with SMM in a reasonable timeframe. We believe this is due to the larger action space available to SVB compared to SPB, which helps SVB to learn a single optimal policy with SAC faster, but causes too much diversity when using SMM, such that the agent never learns useful goal-reaching behavior. 

\subsection{DIAYN}
\label{sec:DIAYN}

Diversity Is All You Need (DIAYN) from \citet{ICML21} aims to construct a group of skills that are maximally diverse but also discriminable. Using Mutual Information (MI), $I(\cdot; \cdot)$ with skills $Z$, actions $A$ and states $s$ the objective function contains terms to:

\begin{itemize}
    \item Minimize MI, between skills and actions in a given state: $-I(A;Z|S)$
    \item Maximize the entropy of the policy: $\mathcal{H[A|S]}$
    \item Maximize MI between skills and states, so skills control which states are visited: $I(S;Z)$
\end{itemize}

Putting this together as an entropy based objective:

$$
    \begin{aligned}
        \mathcal{F}(\theta) &= I(S;Z) + \mathcal{H}[A|S] - I(A;Z|S) \\
                            &= \mathcal{H}[Z] - \mathcal{H}[Z|S] + \mathcal{H}[A|S, Z] \\
    \end{aligned}
$$

The entropy over the policy ($\mathcal{H}[A|S, Z]$) is included as part of the SAC exploration reward. For the remaining terms we can rewrite the objective as an expectation, replacing $p(z|\mathbf{s}_{t})$ with a discriminator $q_{\phi}(z|\mathbf{s}_{t})$, bounded using Jensen's inequality as a variational lower bound $\mathcal{G}(\phi, \theta)$ on the objective $\mathcal{F}(\theta)$.

$$
    \begin{aligned}
        \mathcal{F}(\theta) &= \mathcal{H}[A|S, Z] + \mathcal{H}[Z] - \mathcal{H}[Z|S] \\
                            &= \mathbb{E}_{z \sim p(z), s \sim \pi(z)}[\mathcal{H}(\pi(\cdot | s))] + \mathbb{E}_{z \sim p(z), s \sim \pi(z)}[\log p(z|s)] - \mathbb{E}_{z \sim p(z)} [\log p(z)] \\
                            &\geq \mathbb{E}_{z \sim p(z), s \sim \pi(z)}[\alpha \mathcal{H}(\pi(\cdot | s)) + \log q_{\phi}(z | s) - \log p(z)] = \mathcal{G}(\theta, \phi)
    \end{aligned}
$$

This can be bought into the SAC policy objective, in equation \ref{eq:sac-reward}, as an intrinsic reward:

$$
    r_{z}(\mathbf{s}_{t}, \mathbf{a}_{t}) = \log q_{\phi}(\mathbf{z}_{t} | \mathbf{s}_{t}) - \log p(\mathbf{z}_{t})
$$

In this work we use a VAE to learn the skill discriminator $q_{\phi}(z | s)$ and a fixed one-hot discrete skill vector representation for $Z \sim p(z)$. $z$ is sampled from $Z$ at the beginning of each episode and is then fixed until the episode terminates. 

\begin{figure}[htb!]
    \centering
    \label{fig:M-diayn_train}
    \includegraphics[width=1.0\textwidth]{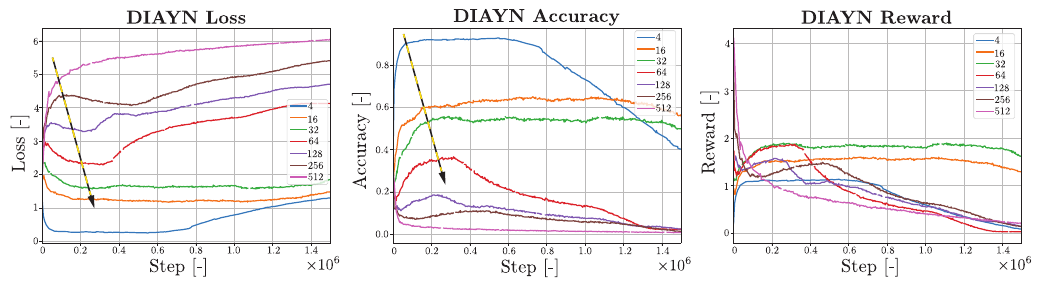}
    \caption{DIAYN training over multiple skill vectors, yellow and black arrow denotes beginning of overfitting.}
\end{figure}

\begin{figure}[htb!]
    \centering
    \label{fig:N-DIAYN_Overfit}
    \includegraphics[width=1.0\textwidth]{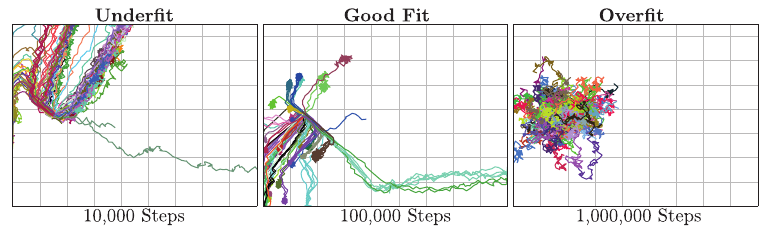}
    \caption{Sampling from DIAYN policy where each colour represents a skill vector $z \in Z$, where $Z$ is a discrete prior represented as a 64 dimension one-hot vector.}
\end{figure}

Figure \ref{fig:M-diayn_train} shows the training plots from DIAYN and how it begins to overfit after too many samples are collected, the effect of this overfitting is shown in figure \ref{fig:N-DIAYN_Overfit}. Note how none of the priors reach the goal-state in the good fit making DIAYN unsuitable for this goal-navigation problem as there are too many diverse states that are not goal-states or useful behaviors. 

\subsection{SMM}
\label{sec:SMM}

State Marginal Matching (SMM) by \citet{ICML22} aims to find a policy $\pi \in \Pi$ that matches the target state density $p^{*}(s)$ to the policy state distribution $\rho_{\pi}(s)$. This objective can be written as:

$$
    \begin{aligned}
        & \min_{\pi \in \Pi} D_{\text{KL}}(\rho_{\pi}(\mathbf{s}_{t}) || p^{*}(\mathbf{s}_{t})) \\
       =& \max_{\pi \in \Pi} \mathbb{E}_{\rho_{\pi(\mathbf{s}_{t})}}[ \log p^{*}(\mathbf{s}_{t}) + \mathcal{H}(\pi(\cdot | \mathbf{s}_{t}))]
    \end{aligned}
$$

The $\rho_{\pi}(s)$ term can be decomposed into a set of policies conditioned on a latent variable $z \in \mathcal{Z}$, the state marginal of this mixture of policies with prior $p(z)$ is then given as:

$$
    \rho_{\pi}(s) = \int_{\mathcal{Z}} \rho_{\pi_{z}}(s) p(z) dz = \mathbb{E}_{z \sim p(z)} [\rho_{\pi_{z}}(s)]
$$

Using Bayes rule and the KL minimization objective the optimization problem becomes:

$$
    \max_{\pi_{z}, z \in \mathcal{Z}} \mathbb{E}_{p(z), \rho_{\pi_{z}}(s)} [r_{z}(s)]
$$

with $r_{z}(s)$:

$$
    r_{z}(s) = \log p^{*}(s) - \log \rho_{\pi_{z}(s)} + \log p(z|s) - \log p(z)
$$

The $\log p(z|s) - \log p(z)$ objectives follow from the earlier DIAYN objective formulation, where we aim to identify distinguishable behaviors whilst exploring the state-space. The $\log p^{*}(s)$ is simply the target distribution and $\log \rho_{\pi_{z}(s)}$ term aims to explore unvisited states.

\begin{table}[htb!]
    \centering
    \caption{State Marginal Matching (SMM) Parameters}
    \label{Tab:smm-params}
    \begin{tabular}{ll}
    \hline
    \textbf{SMM Parameter}                                        & \textbf{Value} \\ \hline
    Skill Distribution                                            & Discrete        \\
    Skill Number                                                  & 4              \\
    Learning Rate Discriminator ($\lambda_{\text{sr}}$)                  & $10^{-3}$      \\
    Learning Rate VAE ($\lambda_{\text{vae}}$)                           & $10^{-2}$      \\
    VAE Beta $(\beta)$                                            & 0.5            \\
    State Entropy Coefficient (${c}_{\mathcal{H}(S|Z)}$)          & 0.1           \\
    Latent Entropy Coefficient ($c_{\mathcal{H}(Z)}$)               & 0.1            \\
    Latent Conditional Entropy Coefficient ($c_{\mathcal{H}(Z|S)}$) & 0.1          \\ \hline 
    \end{tabular}
\end{table}

Table \ref{Tab:smm-params} describes the parameters used for training SMM. We found it necessary to reduce the size of each of the entropy related terms to ensure progress towards the $p^{*}$ objective.

\subsection{APS \& CIC}
\label{sec:APS}

We considered the use of both CIC \citep{ICML40} and APS \citep{ICML23} for the experiments in appendix \ref{sec:te_and_c}. Whilst both offer greater sample efficiency than DIAYN they do not allow for the $p^{*}$ term used by SMM. We found this term necessary to achieve usable datasets within the bounds of the compute we have available.

APS is similar to DIAYN but uses successor features to estimate the MI decomposition:
\begin{itemize}
    \item $\mathcal{H}[Z]$ is estimated using the particle estimator described in APT \citep{ICML43, ICML44}
    \item $\mathcal{H}[Z|S]$ is estimated with successor features, described in VISR \citep{ICML45}
\end{itemize}

Similarly, CIC uses the MI decomposition with Contrastive Predictive Control (CPC) \citep{ICML46} to learn a prior encoded policy. In future work it would be interesting to investigate if the improved discriminator presented in CIC can be used with the same state marginal matching objective found in SMM to create a more effective discriminator in more complex state spaces.  


\end{document}